\PassOptionsToPackage{table,dvipsnames}{xcolor}

\documentclass[10pt,twocolumn,letterpaper]{article}

\usepackage[pagenumbers]{iccv} 

\newcommand \footnoteONLYtext[1]
{
    \let \mybackup \thefootnote
    \let \thefootnote \relax
    \footnotetext{#1}
    \let \thefootnote \mybackup
    \let \mybackup \imareallyundefinedcommand
}

\usepackage{stfloats}
\usepackage{multirow}
\usepackage{longtable}
\usepackage{wrapfig}

\usepackage{soul}
\usepackage[table]{xcolor}         
\definecolor{darkgreen}{RGB}{84, 130, 53}
\definecolor{darkred}{RGB}{192, 0, 0}
\definecolor{myhighlightcolor_gray}{RGB}{229,229,229}
\definecolor{myhighlightcolor_brown}{RGB}{245,237,230}
\usepackage{wrapfig}

\usepackage{amsmath}
\usepackage[most]{tcolorbox}
\usepackage{makecell}
\usepackage{changepage}

\definecolor{iccvblue}{rgb}{0.21,0.49,0.74}
\usepackage[pagebackref,breaklinks,colorlinks,allcolors=iccvblue]{hyperref}

\def\paperID{10618} 
\def\confName{ICCV}
\def\confYear{2025}

\title{FineMotion: A Dataset and Benchmark with both Spatial and Temporal Annotation for Fine-grained Motion Generation and Editing}

\author{
Bizhu Wu$^{1,2,4}$\quad Jinheng Xie$^{5}$\quad Meidan Ding$^{1,4}$\quad Zhe Kong$^6$\quad Jianfeng Ren$^{2}$\thanks{Corresponding authors}\quad Ruibin Bai$^{2}$\quad \\ 
Rong Qu$^7$\quad Linlin Shen$^{2,3,4*}$ \\
$^1$ School of Computer Science \& Software Engineering, Shenzhen University \\ 
$^2$ School of Computer Science, University of Nottingham Ningbo China, Ningbo, China \\ 
$^3$ Computer Vision Institute, School of Artificial Intelligence, Shenzhen University \\
$^4$ Guangdong Provincial Key Laboratory of Intelligent Information Processing \\
$^5$ National University of Singapore \quad $^6$ Shenzhen Campus of Sun Yat-sen University \\ 
$^7$ School of Computer Science, University of Nottingham, Nottingham, United Kingdom \\
{\tt\small wubizhu@email.szu.edu.cn, jianfeng.ren@nottingham.edu.cn, llshen@szu.edu.cn} \\ 
}

\begin{document}

\twocolumn[{%
\renewcommand\twocolumn[1][]{#1}%
\maketitle
\begin{center}
    \centering
    \captionsetup{type=figure}
    \vspace{-2em}
    \setlength{\abovecaptionskip}{3pt}  
    \setlength{\belowcaptionskip}{0pt}  
    \includegraphics[width=1.0\linewidth]{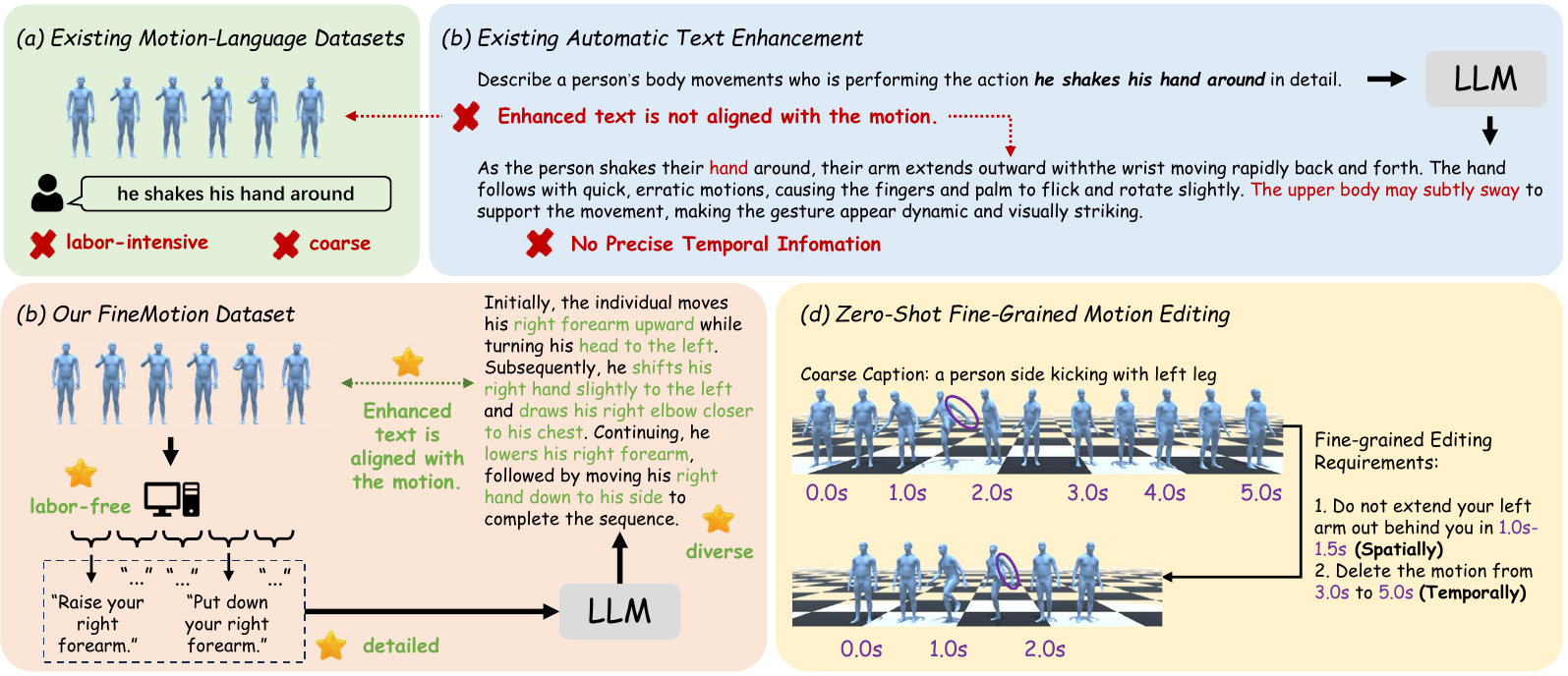}
    \captionof{figure}{
        Illustration of 
        \textbf{(a) Existing Motion-Language Datasets} are manually annotated, with textual descriptions that are coarse and lack detail.
        \textbf{(b) Existing textual enhancement} works obtained more detailed descriptions of a motion phrase or caption via large language models, but failed to align with the actual motion sequence.
        \textbf{(c) Our FineMotion dataset} features strictly aligned and fine-grained descriptions of human body part movements for both motion snippets (short segments of motion sequences) and entire motion sequences, while being easily scalable.
        \textbf{(d)} The proposed dataset further enables \textbf{zero-shot fine-grained motion editing} capabilities.
    }
    \label{fig:application}
    \vspace{1.0em}
\end{center}%
}]
\footnoteONLYtext{*Corresponding Author}


\begin{abstract}
Generating realistic human motions from textual descriptions has undergone significant advancements. 
However, existing methods often overlook specific body part movements and their timing.
In this paper, we address this issue by enriching the textual description with more details.
Specifically, we propose the FineMotion dataset, which contains over 442,000 human motion snippets — short segments of human motion sequences — and their corresponding detailed descriptions of human body part movements.
Additionally, the dataset includes about 95k detailed paragraphs describing the movements of human body parts of entire motion sequences.
Experimental results demonstrate the significance of our dataset on the text-driven fine-grained human motion generation task, especially with a remarkable +15.3\% improvement in Top-3 accuracy for the MDM model.
Notably, we further support a zero-shot pipeline of fine-grained motion editing, which focuses on detailed editing in both spatial and temporal dimensions via text.
Dataset and code available at: \href{https://github.com/CVI-SZU/FineMotion}{CVI-SZU/FineMotion}  
\end{abstract}


\section{Introduction}
\label{sec:intro}

The text-to-motion task involves generating human motion sequences from natural language textual descriptions, and has attracted growing interest due to its applications in animation making, virtual reality, and robotics.
Recent advancements in this task~\cite{mdm, petrovich2022temos, t2mgpt, humanml3d, kim2023flame} have advanced its practical deployment.
Nowadays, people have higher expectations for this task, emphasizing the need to be controllable and realistic.

\textit{"A man waves his right hand"} is an example textual description from the popular text-motion pair dataset, HumanML3D~\cite{humanml3d}. 
The motion sequences generated from this description may contain three stages: raising the right hand near the head, waving it, and lowering it.
However, several questions arise:
(1) \textit{When should the right hand be raised or lowered?}
(2) \textit{How long should the wave last?}
(3) \textit{Are other body parts involved?}
Obviously, textual descriptions in existing text-motion pair datasets, like HumanML3D~\cite{humanml3d} and KIT-ML~\cite{kitml}, are \textbf{too coarse and less informative} to answer these questions.

To address these limitations, several works~\cite{kalakonda2023actiongpt, athanasiousinc} have proposed extending existing coarse captions with detailed descriptions using Large Language Models (LLMs).
Since current LLMs cannot directly process motion sequences to generate corresponding detailed descriptions, these works rely on language-only LLMs.
They prompt LLMs with only coarse captions as input to generate detailed body part movement descriptions based on LLMs' own biases, as illustrated in Fig.~\ref{fig:application}(b).
However, it is evident that the enhanced textual output \textbf{fails to precisely align with the actual human motion sequences}.

\begin{table*}[!b]
\begin{center}
\footnotesize
\vspace{-0.5em}
\setlength{\tabcolsep}{5pt} 
\begin{tabular}{cccccccc}
    \toprule[1pt]
    
    Dataset & Year & Number of Motions  & Number of Texts & Granularity & Annotation Source & Easily Scalable \\ 
    
    \midrule[0.5pt]

    KIT-ML~\cite{kitml} & 2016 & 3,911 & 6,278 & Coarse & Human & $\times$ \\
    
    HumanML3D~\cite{humanml3d} & 2022 & 14,616 & 44,970 & Coarse & Human & $\times$ \\

    HuMMan-MoGen~\cite{zhang2024finemogen} & 2023 & 2,968 & 102,336 & Fine & Human & $\times$ \\
    
    \textbf{FineMotion (Ours)} & 2024 & 14,616 & 442,314 (Snippet) + 94,432 (Sequence) & Fine & Auto + Human & \checkmark \\

    \bottomrule[1pt]
    
\end{tabular}
\setlength{\abovecaptionskip}{5pt} 
\caption{
    Comparisons of 3D human motion-language datasets.
} \label{table: dataset_comparison}
\end{center}
\end{table*}

\begin{figure*}[!t]
\setlength{\abovecaptionskip}{3pt}  
\includegraphics[width=1.0\linewidth]{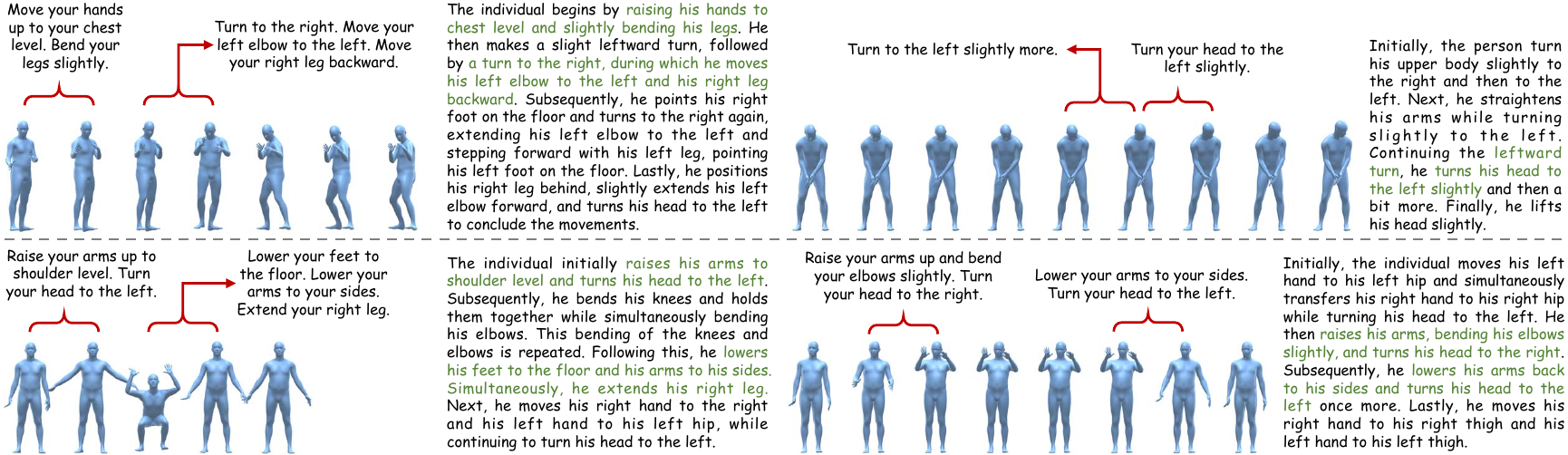}
    \caption{
        \textbf{
        Examples from the FineMotion dataset.}
        \textbf{Top}: Human-annotated BPM snippet descriptions and paragraphs.
        \textbf{Bottom}: Automatically generated BPM snippet descriptions and paragraphs. 
        \textcolor{darkgreen}{Colored text} in paragraphs links to the corresponding snippet descriptions.
    }
    \label{fig:dataset_examples}
\vspace{-0.8em}
\end{figure*}

In this work, we construct a new dataset, \textbf{FineMotion}, which provides precise \underline{B}ody \underline{P}art \underline{M}ovement (BPM) descriptions for short temporal intervals in human motion sequences. 
It contains about 420k automatically generated BPM descriptions for motion snippets (short segments of motion sequences), referred to as \textbf{BPMSD}, and over 21k human-annotated ones. 
Additionally, it includes around 95k BPM Paragraph (\textbf{BPMP}) for detailed descriptions of entire human motion sequences.
Examples are shown in Fig.~\ref{fig:dataset_examples}.
Notably, the temporal information embedded in the textual annotations allows for easy augmentation, such as random cropping along the temporal dimension to generate numerous pairs of motion clips (composed of several adjacent snippets) and their corresponding BPM descriptions.

Tab.~\ref{table: dataset_comparison} compares our FineMotion with existing text-motion pair datasets.
The textual descriptions in the KIT-ML\cite{kitml} and HumanML3D~\cite{humanml3d} datasets are coarse and lack detail.
HuMMan-MoGen~\cite{zhang2024finemogen} provides detailed body part movement descriptions for each phase, but relies on manually specifying the start and end points of standardized phases, limiting scalability.
In contrast, we not only include fine-grained textual descriptions for motions in our FineMotion dataset, but also propose an efficient and scalable pipeline for the automatic generation of detailed textual descriptions, facilitating easy dataset expansion.

With FineMotion, we establish a benchmark to evaluate several state-of-the-art motion generation methods. 
Comprehensive experiments demonstrate its effectiveness in producing precise and realistic motion. 
Building on this foundation, we further explore a zero-shot fine-grained motion editing pipeline, enabling users to modify descriptions to adjust motion content. 
This improves interaction efficiency and broadens applications.

Overall, our key contributions are as follows:
\textbf{First}, we develop an efficient and scalable pipeline for the automatic generation of detailed motion descriptions. 
It offers a potent solution to the imprecision and lack of specificity issues in the motion description annotations domain.
\textbf{Second}, the proposed FineMotion dataset bridges the domain gap with over 442k textual descriptions for short motion snippets, and around 95k paragraphs for whole motion sequences, all informative and strictly aligned.
\textbf{Third}, we validate the dataset’s effectiveness and generalization by benchmarking classical text-to-motion models that are intricately adapted and carefully tailored to handle our long and detailed text. 
Experimental results show that all these models exhibit notable performance gains, particularly with a +15.3\% increase in Top-3 retrieval accuracy for our MDM variant.
\textbf{Finally}, we demonstrate a zero-shot fine-grained motion editing pipeline, enabling controllable and realistic motion generation via textual modifications.


\section{Related Work}
\label{sec:related_work}

\textbf{Text-Driven Human Motion Generation.}
Motion generation can be generated from various conditions, including text~\cite{t2mgpt, zhang2024motiongpt, mdm, petrovich2022temos, kim2023flame}, action classes~\cite{actor, raab2023modi, tevet2022motionclip, kalakonda2023actiongpt}, and music~\cite{li2022danceformer, li2024exploring}.
Among these, text-to-motion stands out for its user-friendly language interface.
MotionDiffuse~\cite{zhang2024motiondiffuse} and MDM~\cite{mdm} integrated diffusion models for text-to-motion generation with impressive results. 
T2M-GPT~\cite{t2mgpt} and MoMask~\cite{guo2024momask} quantized human motions and used transformer networks to generate high-quality motion. 
Recently, MotionGPT~\cite{openmotionlab_motiongpt} treated motion as a foreign language in a natural language model.
However, these methods rely on coarse captions that overlook fine-grained action details.
In contrast, we focus on detailed textual descriptions that capture finer body part movements over time, helping generate motions more similar to the ground truth.

\vspace{0.5em}
\noindent\textbf{Human Motion and Language Data.}
Existing datasets like KIT-ML~\cite{kitml} and HumanML3D~\cite{humanml3d} offer textual annotations for 3D motions but lack fine-grained details.
To address this, some works leveraged large language models for enhanced semantic annotations.  
Action-GPT~\cite{kalakonda2023actiongpt} used carefully designed prompts to generate more detailed descriptions, but these may not align well with ground-truth motion. 
SemanticBoost~\cite{he2023semanticboost} and MotionScript~\cite{yazdian2023motionscript} mapped body part movements to predefined statuses but ignored precise temporal information and relied on fixed templates.
FineMoGen~\cite{zhang2024finemogen} introduced the HuMMan-MoGen dataset with fine-grained spatio-temporal descriptions, but required extensive manual annotation of temporal boundaries, limiting scalability.
Building on prior work, we present FineMotion, a new dataset with detailed, temporally precise, diverse, and motion-aligned annotations. 
Besides, our automatic dataset construction pipeline facilitates easy scaling.

\vspace{0.5em}
\noindent\textbf{Text-Driven Human Motion Editing.}
Several text-to-motion approaches~\cite{kim2023flame, mdm, tevet2022motionclip, tseng2023edge} have explored human motion editing.
Diffusion model-based methods~\cite{kim2023flame, mdm, tseng2023edge} diffused a reference motion, masked specific frames and joints, replaced the masked parts with the ones conditioned on another coarse text, and denoised to obtain the edited human motion.
MotionCLIP~\cite{tevet2022motionclip} performs editing via latent space arithmetic.
However, these methods rely on coarse descriptions, which limit their control over specific body parts, timing, and duration.
In contrast, our baseline models generate motion strictly based on detailed descriptions, facilitating fine-grained motion editing across temporal and spatial dimensions through precise text editing.


\section{The proposed FineMotion dataset}
\label{sec: finemotion_dataset}

The proposed FineMotion dataset builds upon HumanML3D~\cite{humanml3d} by describing motions in fine details both spatially and temporally. The motion sequences, sourced from AMASS~\cite{mahmood2019amass} and HumanAct12~\cite{humanact12}, span diverse actions like `walking’, `swimming’, and `dancing’. 
They are pre-processed by scaling to 20 FPS, randomly cropping those longer than 10 seconds, re-targeting to a standard skeletal template, and rotating to face the Z+ direction.

As for the textual descriptions, we include two types: 
One is the body part movement description for the motion snippet, a short segment from the motion sequence, short for \textbf{BPMSD};
The other one is the body part movement description paragraph for the whole motion sequence, short for \textbf{BPMP}.
Generally, the enriched textual descriptions have the following three properties:
(1) \textit{more fine-grained descriptions of body part movements}, 
(2) \textit{precise temporal information}, and 
(3) \textit{more diverse}. 
We next present the dataset construction pipeline and some dataset statistics.

\subsection{Dataset Construction Pipeline}
\label{sec:pipeline}

\begin{figure}[!t]
\begin{center}
\includegraphics[width=1.0\linewidth]{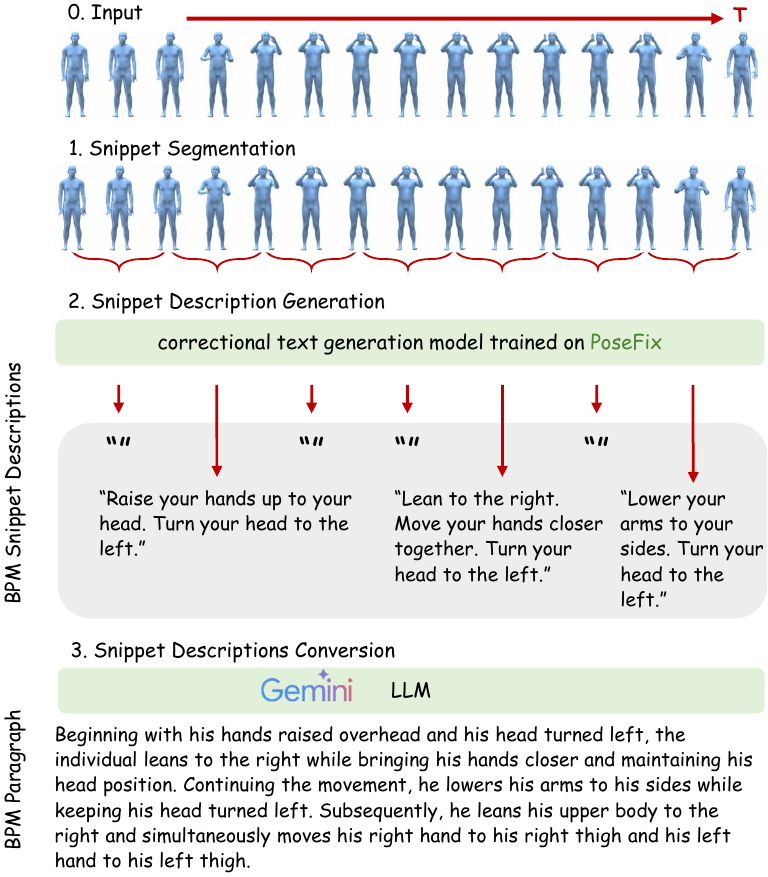}
\end{center}
\setlength{\abovecaptionskip}{-0.5em}  
\setlength{\belowcaptionskip}{0pt}  
   \caption{
        The construction pipeline of our FineMotion dataset.
   }
    \label{fig:dataset_pipeline}
\vspace{-0.8em}
\end{figure}

The pipeline for enhancing text descriptions from a human motion sequence is illustrated in Fig.~\ref{fig:dataset_pipeline}. 
The input is the SMPL~\cite{loper2023smpl} pose parameters of the human motion sequence, 
while the pipeline outputs two types of BPM descriptions over time in English. \textit{i.e.}, BPM Snippet Description and BPM Paragraph. 
The process comprises three steps: 
\begin{enumerate}[left=0pt, labelindent=0pt]
    \item Given a motion sequence, we divide it into short snippets along the temporal dimension  (Sec.~\ref{sec:SG}).

    \item The detailed BPM descriptions are generated for each snippet (Sec.~\ref{sec:SDG}).
    
    \item All snippet descriptions from the same motion sequence are further organized into a paragraph (Sec.~\ref{sec:PG}).
\end{enumerate}
Notably, this pipeline is \textbf{universal} and can be applied to any text-motion pair datasets or motion-only datasets.

\subsubsection{Snippet Segmentation}
\label{sec:SG}

To obtain detailed, temporally aligned descriptions of motions, we first explicitly segment each motion into short snippets along the temporal dimension.

In this dataset, we choose to fix the snippet duration for two main reasons:
First, a fixed snippet duration simplifies the dataset scaling process within our automated dataset construction pipeline. 
It \textbf{minimizes the need for manual annotation to determine the start and end points of each snippet}. 
Secondly, a consistent duration reduces the complexity of the fine-grained motion generation model, as it \textbf{eliminates the need for additional inputs, such as the start and end points for each snippet}. 
Otherwise, the model would require these inputs to be explicitly aligned with each snippet's fine-grained description.

\begin{figure}[!t]
\begin{center}
\includegraphics[width=1.0\linewidth]{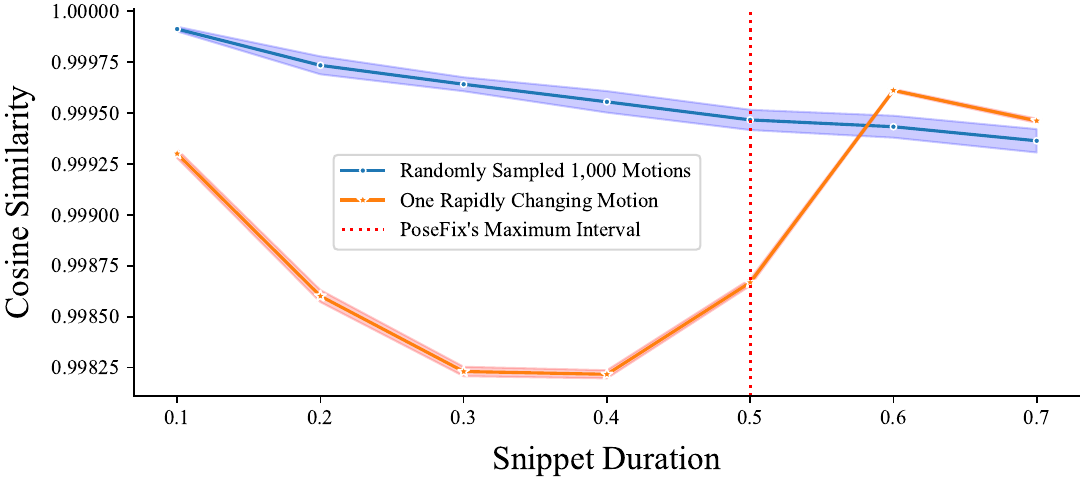}
\end{center}
\setlength{\abovecaptionskip}{-0.5em}  
   \caption{
        Mean and 95\% confidence interval of the cosine similarity between semantic features of snippets' start and end poses.
   }
    \label{fig:statistics_snippet}
\vspace{-0.8em}
\end{figure}

To determine the optimal snippet duration $T_s$, we propose two guiding principles to help researchers tailor this value to their own datasets:
First, \textbf{select $T_s$ to minimize similarity between snippets' start and end poses}.
As shown in Fig.~\ref{fig:statistics_snippet}, we calculate cosine similarity between PoseScript~\cite{delmas2022posescript} semantic features of snippets' start and end poses.
The analysis was performed on randomly sampled snippets from all motions in our dataset (\textit{blue} line) and a representative example of a rapidly changing motion (\textit{orange} line), with various durations. 
Results show that motions in our dataset generally progress slowly, suggesting that a longer interval helps reduce redundancy among snippet descriptions. 
Further details are available in the appendix.
Meanwhile, PoseFix~\cite{delmas2023posefix} suggests that larger time differences between two poses allow for a wide range of plausible in-between motions.
Therefore, the second principle is that \textbf{the value of $T_s$ should not exceed 0.5s}, which is the maximum time difference for pose pair selection specified by PoseFix~\cite{delmas2023posefix}.
Following these principles, we set $T_s$ to 0.5s.
Notably, any remaining segment of a motion sequence shorter than $T_s$ is also treated as an individual snippet.

\subsubsection{Snippet Description Generation}
\label{sec:SDG}

\textbf{Collection of Automatically Generated Annotations.}
This step builds upon the outstanding foundation work, PoseFix~\cite{delmas2023posefix}.
It developed a correctional text generation model to describe how body parts in a source pose should be modified to achieve a target pose.
Specifically, the model integrates two pose embeddings in the cross-attention mechanisms of a text transformer to generate correctional text for a given pose pair.
We believe that the correctional text describing the transition between start and end poses of a snippet effectively captures the body part movements within it, and can thus be naturally regarded as the detailed BPM description for this snippet.

\vspace{0.5em}
\noindent\textbf{Collection of Human Annotations.}
We recruited eight undergraduate students with strong English comprehension skills to manually annotate motion sequences. 
Specifically, we first constructed a BPM description corpus, which is composed of sentences derived from the automatically generated annotations for all snippets.
Then, for each motion sequence, annotators were provided with its automatically generated annotations to ease their workload.
If those annotations were inappropriate, they were instructed to select suitable sentences from the corpus. 
This ensured that the manually annotated descriptions closely match the style of the automatically generated ones. 
Besides, annotators were required to remove trivial or passive body part movements (such as `\textit{move your left leg backward}' in a `\textit{walking forward}' motion sequence) from each snippet description. 
For snippets involving rapid changes (\textit{i.e.}, containing multiple stages), annotators were encouraged to describe each stage separately and use phrases such as “\textit{Then,}” to connect them.

After the preliminary annotations, two additional well-trained students conducted two rounds of manual checks to scrutinize, correct, and supplement the content.
To fix any remaining grammar and spelling issues, we utilized Grammarly~\cite{grammarly}. 
Overall, we annotated around 21k motion snippets (\textit{i.e.}, 5\% of the motion sequences).
The upper part of Fig.~\ref{fig:dataset_examples} shows examples of motion sequences with their manually annotated BPM snippet descriptions; more can be found in the appendix.

\subsubsection{Paragraph Generation}
\label{sec:PG}

Apart from the detailed BPM description for the motion snippet, we also include that for the whole motion sequence in the FineMotion dataset, referred to as the BPM paragraph.
Instead of applying fixed templates to connect the descriptions for all the snippets in a single human motion sequence, we leverage Gemini’s~\cite{team2023gemini} advanced language capabilities to create diverse, coherent paragraphs that follow natural human logic.

Specifically, for each human motion sequence, we first remove empty BPM snippet descriptions, and connect the rest with numbers to preserve temporal order, resulting in \textcolor{darkgreen}{\underline{\textit{BPMSDs}}} (see example in \textcolor{magenta}{\#\#\# Input \#\#\#}).
Following ActionGPT~\cite{kalakonda2023actiongpt}, we craft a prompt to guide Gemini to organize these descriptions.
After multiple trials, we determine the following prompt.
It first introduces the task, \textit{i.e.}, arranging all snippet descriptions into a cohesive paragraph. 
Then, it lists several requirements to ensure continuity, proper time order, converting PoseFix's imperative output to descriptive ones, reliability, and completeness for the output paragraph. 
An example is also included to help Gemini understand and perform the conversion effectively.

\begin{tcolorbox}[breakable, boxrule=0pt, colframe=white, sharp corners, left=1mm, right=1mm, top=1mm, bottom=1mm]

\begin{scriptsize}

    Given a list of items describe the body part's movements that happen in succession. Note that each item describes the body part's movements that happen simultaneously. The items are already arranged in chronological order. Please organize these items into a single paragraph according to the following requirements:

    \vspace{0.5em}
    
    (1) If some of the body part's movements in adjacent two items are similar, then use vocabulary indicating their continuity, such as `keep' and `continue'. 
    
    (2) Use proper conjunctions of time to connect all items but do not add too many additional descriptions. 
    
    (3) Change the output into descriptive statements. For example, the subject can be changed to `The person', `The individual', etc.
    
    (4) Make sure not to add any extra body part's movement that does not exist in the original list of items.
    
    (5) Make sure all the body part's movements in the list are described in the output.

    \vspace{0.5em}
    
    \textcolor{magenta}{\#\#\# Input \#\#\#}: 1. Lower your left leg, keeping the point at your foot and curving it very slightly. Bring down your right hand even to your left hand.
    2. Raise your right leg. Bring down your right hand.
    3. Lower your legs so your feet and upper legs are parallel with the ground. Bend your right arm slightly to have it hover over your elevated legs. Lower your left forearm so it is at an angle pointing down.

    \vspace{0.5em}
    
    \#\#\# Output \#\#\#: The individual lowers his left leg, ensuring the foot is pointed and slightly curved, while simultaneously bringing his right hand down to meet the left. Afterward, he raises his right leg and continues bringing the right hand down. Consequently, he lowers his legs until the feet and upper legs are parallel to the floor. At the same time, he slightly bends his right arm, positioning it above his elevated legs, and lowers his left forearm, forming a downward angle.

    \vspace{0.5em}
    
    \#\#\# Input \#\#\#:
    [\textcolor{darkgreen}{\underline{\textit{BPMSDs}}}]

    \vspace{0.5em}
    
    \#\#\# Output \#\#\#:

\end{scriptsize}
    
\end{tcolorbox}

One should notice that the input to Gemini here is all BPM snippet descriptions within a motion sequence. 
All Gemini has to do is connect these precise snippet descriptions into a coherent paragraph. 
As a result, the generated BPM paragraphs are \textbf{strictly aligned with the actual motion sequence}, which addresses the limitations mentioned in Sec.~\ref{sec:intro}. 
Interestingly, instead of merely connecting all the snippet descriptions, Gemini sometimes offers precise paraphrases, further \textbf{enhancing the diversity} of BPM paragraphs.
Examples can refer to Fig.~\ref{fig:dataset_examples} and the appendix.

\subsection{Statistics Analysis}
\label{sec:statistics}

\begin{figure}[!b]
\vspace{-0.5em}
\begin{center}
\includegraphics[width=1.0\linewidth]{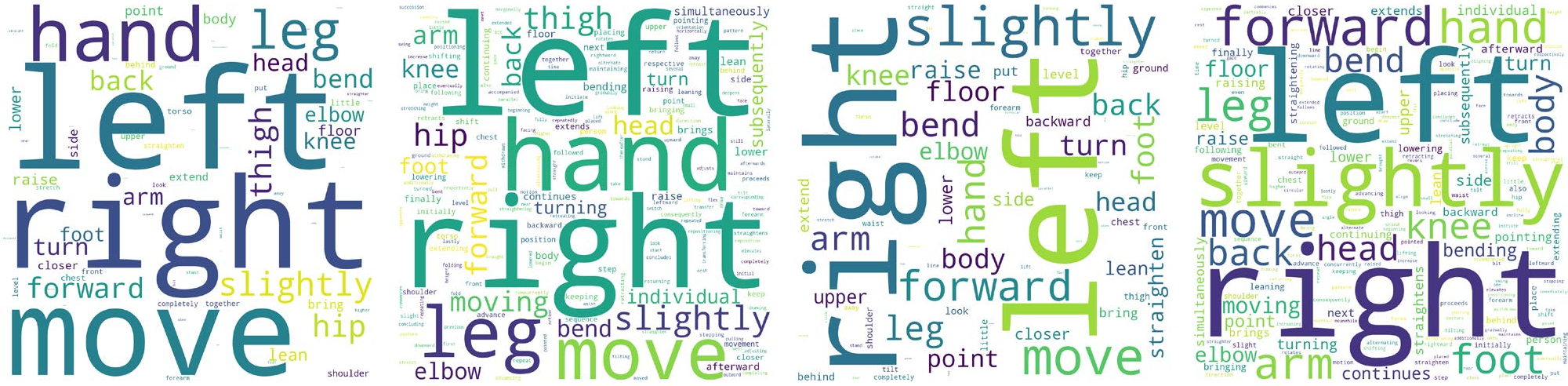}
\end{center}
\vspace{-1.5em}
   \caption{
        \textbf{Visualizations of the 200 most frequent words in our textual descriptions.} 
        From left to right are the word clouds of automatically-generated BPMSD, automatically-generated BPMP, human-annotated BPMSD, and human-annotated BPMP.
   }
    \label{fig:wordcloud}
\end{figure}

In summary, our FineMotion dataset contains 21,346 human-annotated BPM descriptions and 420,968 automatically generated ones for diverse motion snippets, \textit{i.e.}, \textbf{BPMSD}.
Notably, the temporal information within the textual annotations facilitates \textbf{easy augmentation}, such as performing random cropping along the temporal dimension. 
This approach can generate numerous pairs of motion clips—each consisting of several adjacent snippets—along with their corresponding BPM descriptions.
Additionally, the dataset includes 4,492 BPM paragraphs (\textbf{BPMP}) organized from human-annotated BPM snippet descriptions,
and 89,940 paragraphs organized from automatically generated BPM snippet descriptions, 
covering a total of 29,232 motion sequences.

For BPMSD, the average and median lengths are 18 and 19 words, respectively.
As for BPMP, the average and median lengths are 247 and 221 words, respectively. 
In terms of \textbf{data splits}, we adopt the same ones as HumanML3D, in which the data is partitioned into 80\%, 5\%, and 15\% for training, validation, and testing, respectively, ensuring no overlap among the subsets.
Fig.~\ref{fig:wordcloud} shows the most frequent words in the FineMotion dataset.
Thanks to the assistance of Gemini, the vocabulary in BPMP (the second and fourth instances) is notably more diverse.


\section{Experiments}
\label{sec:Exp}

In this section, we first validate the accuracy of our textual annotation pipeline.
Then, we build a text-driven fine-grained human motion generation benchmark on FineMotion.
Finally, we explain how we have implemented zero-shot fine-grained motion editing with our dataset.

\subsection{Data Preprocessing}
\label{sec:text_process}

We follow~\cite{humanml3d} to pre-process motion sequences into $d = 263$-dimensional features.
For textual input, we use coarse descriptions in HumanML3D~\cite{humanml3d} (denoted as `\textbf{T}'), 
and the detailed BPM snippet descriptions (BPMSDs) or BPM paragraph (BPMP) from our FineMotion dataset (denoted as `\textbf{DT}').
Notably, since a BPMSD covers only a short interval rather than the entire sequence, we use a fixed template to connect all BPMSDs in a motion sequence.
Specifically, empty snippet descriptions, which indicate no significant BPM, are replaced with the special token \textcolor{orange}{\textless Motionless\textgreater}.
We then use the special token \textcolor{cyan}{\textless SEP\textgreater} to connect snippet descriptions across intervals and preserve temporal information, \textit{e.g.},

\vspace{-0.5em}
\begin{tcolorbox}[boxrule=0pt, colframe=white, sharp corners, left=1mm, right=1mm, top=1mm, bottom=1mm]

\begin{scriptsize}
    
    \textcolor{lightgray}{Given all the BPM snippet descriptions from a motion sequence:}   

    ["", "", "Move your right leg forward slightly.", "Turn to the left. Move your left leg forward. Move your left hand back slightly.", "Lean to the right. Move your right leg forward."]

    \vspace{0.5em}
    
    \textcolor{lightgray}{Fit into the template:}
    
    "\textcolor{orange}{\textless Motionless\textgreater} \textcolor{cyan}{\textless SEP\textgreater} \textcolor{orange}{\textless Motionless\textgreater} \textcolor{cyan}{\textless SEP\textgreater} 
    Move your right leg forward slightly. \textcolor{cyan}{\textless SEP\textgreater}
    Turn to the left. Move your left leg forward. Move your left hand back slightly. \textcolor{cyan}{\textless SEP\textgreater}
    Lean to the right. Move your right leg forward."

\end{scriptsize}

\end{tcolorbox}

\begin{table*}[!b]
\begin{center}
\footnotesize
\setlength{\tabcolsep}{6pt} 
\begin{tabular}{cccccccccc}
    \toprule[1pt]
    
    & \multicolumn{3}{c}{Train Set} & \multicolumn{6}{c}{Test Set} \\

    \cmidrule(r){2-4}\cmidrule(r){5-10}

    & \multirow{2}{*}{T2M} & \multirow{2}{*}{(T\&DT$^\text{Auto}$)2M} & \multirow{2}{*}{(T\&DT$^\text{Human}$)2M} & \multicolumn{2}{c}{T2M} & \multicolumn{2}{c}{(T\&DT$^\text{Auto}$)2M} & \multicolumn{2}{c}{(T\&DT$^\text{Human}$)2M} \\
     
    \cmidrule(r){5-6}\cmidrule(r){7-8}\cmidrule(r){9-10}
     
    & & & & R-Top3 $\uparrow$ & FID $\downarrow$ & R-Top3 $\uparrow$ & FID $\downarrow$ & R-Top3 $\uparrow$ & FID $\downarrow$ \\
     
    \midrule[1pt]

    (1) & \checkmark & - & - & 0.781$^{\pm.002}$ & 0.123$^{\pm.005}$ & - & - & - & - \\

    \midrule[1pt]

    \rowcolor{gray!20} \multicolumn{10}{l}{\textit{\textcolor{gray}{DT: BPMSD}}} \\
    
    (2) & \checkmark & \checkmark & - & 0.784$^{\pm.002}$ & 0.124$^{\pm.005}$ & \textbf{0.789$^{\pm.002}$} & \textbf{0.091$^{\pm.003}$} & - & - \\
    
    (3) & \checkmark & \checkmark & \checkmark & 0.781$^{\pm.002}$ & 0.154$^{\pm.007}$ & \textbf{0.789$^{\pm.002}$} & 0.112$^{\pm.005}$ & \textbf{0.789$^{\pm.002}$} & \textbf{0.091$^{\pm.004}$} \\
    
    \midrule[1pt]
    
    \rowcolor{gray!20} \multicolumn{10}{l}{\textit{\textcolor{gray}{DT: BPMP}}} \\

    (4) & \checkmark & \checkmark & - & 0.779$^{\pm.002}$ & 0.136$^{\pm.005}$ & 0.785$^{\pm.002}$ & 0.102$^{\pm.004}$& - & - \\
    
    (5) & \checkmark & \checkmark & \checkmark & 0.781$^{\pm.002}$ & 0.155$^{\pm.006}$ & \textbf{0.788$^{\pm.002}$} & 0.104$^{\pm.005}$ & \textbf{0.788$^{\pm.002}$} & \textbf{0.100$^{\pm.005}$} \\

    \bottomrule[1pt]
    
\end{tabular}
\setlength{\abovecaptionskip}{5pt} 
\caption{
    \textbf{Evaluation of our textual annotation pipeline with (T\&DT)2M-GPT.}
    `T' means coarse descriptions on the HumanML3D, while `DT' means detailed texts on our FineMotion dataset.
    We repeat all evaluations 20 times and report the average with a 95\% confidence interval.
    \textbf{Bold} text means the best results in each block.
    Results show that incorporating our fine-grained and human-annotated texts enhances motion generation performance, which proves the quality of our textual annotation pipeline.
} \label{table: ablation}
\end{center}
\end{table*}

\subsection{Evaluation Metrics} 

We evaluate the generated motions using metrics from~\cite{t2mgpt,humanml3d}: Frechet Inception Distance (FID), Multi-modal Distance (MM-Dist), R-Precision Top-1/2/3, Diversity, and Multi-modality (MModality). 
These metrics assess the realism and diversity of synthesized motions, with definitions provided in~\cite{t2mgpt,humanml3d}.


\subsection{Baseline Models}
To accommodate both coarse and detailed descriptions, we intricately adapted three classical text-to-motion models, \textit{i.e.}, MDM~\cite{mdm}, T2M-GPT~\cite{t2mgpt}, and MoMask~\cite{guo2024momask}, to better handle our long, detailed text.
Specifically, we replace the CLIP~\cite{radford2021clip} text encoder with T5-Base~\cite{t5} to avoid truncation of over-length detailed text.
We then apply mean pooling along the sequence length dimension of the T5-Base encoder output to obtain a single text embedding.
Notably, instead of directly connecting coarse captions and detailed text into one before encoding, our variants are trained to synthesize motion from the concatenated embeddings of both components.
We refer to these adapted variants as (T\&DT)2M-MDM, (T\&DT)2M-GPT, and (T\&DT)2M-MoMask, respectively.
More details on model design and ablation studies can be found in the appendix.


\subsection{Evaluation of the Textual Annotation Pipeline} 

We validate the quality of our detailed text using our (T\&DT)2M-GPT variant in Tab.~\ref{table: ablation}, where all models are trained under the same setting. 
For fair comparisons, we re-implement T2M-GPT with a T5-Base text encoder as the \textit{baseline} for the coarse-grained text-to-motion (T2M) task.
Baseline results are reported in Tab.~\ref{table: ablation}.(1).
Additional results for other variants are provided in the appendix.

\vspace{0.5em}
\noindent\textbf{Automatically Generated Annotations.} 
We train (T\&DT)2M-GPT on both the T2M task and fine-grained text-to-motion task, denoted as (T\&DT)2M.
Here, DT is automatically generated descriptions (\textbf{DT$^\text{Auto}$}).
When training the T2M task, DT is replaced by the special token \textless EMPTY\textgreater.
From Tab.~\ref{table: ablation}.(2),
(T\&DT)2M-GPT achieve performance with an FID of 0.091 (\textit{vs.} 0.123 of \textit{baseline}) and R-Top3 of 0.789 (\textit{vs.} 0.781 of \textit{baseline}).
Similar gains are observed with our automatically generated BPMP in Tab.\ref{table: ablation}.(4).
These results demonstrate that FineMotion's detailed BPM texts help generate motions more aligned with ground-truth.
Furthermore, the BERTScore between human annotations and automatically generated ones is 0.89, comparable to the scores achieved by translation models such as Transformer-big on WMT14 En-De (0.86) and En-Fr (0.89)~\cite{zhang2019bertscore}.
These pieces of evidence prove the effectiveness and quality of our automatically generated annotations.

\begin{table*}[!t]
\begin{center}
\footnotesize
\setlength{\tabcolsep}{3pt} 
\begin{tabular}{lccccccccc}
    \toprule[1pt]
    
    \multirow{2}{*}{Methods} & \multicolumn{2}{c}{Text Granularity} & \multicolumn{3}{c}{R-Precision $\uparrow$} & \multirow{2}{*}{FID $\downarrow$} & \multirow{2}{*}{MM-Dist $\downarrow$}  & \multirow{2}{*}{Diversity $\rightarrow$}  & \multirow{2}{*}{MModality $\uparrow$} \\ 

    \cmidrule(r){2-3}\cmidrule(r){4-6}

     & Coarse (T) & Fine (DT) & Top-1 & Top-2 & Top-3 & &  \\

    \midrule[0.5pt]

    Real motion & \checkmark & - & 0.511$^{\pm.003}$ & 0.703$^{\pm.003}$ & 0.797$^{\pm.002}$ & 0.002$^{\pm.000}$ & 2.974$^{\pm.008}$ & 9.503$^{\pm.065}$ & - \\
    
    \midrule[0.5pt]
    
    TEMOS~\cite{petrovich2022temos} $_{\text{ECCV'22}}$ & \checkmark & - & 0.424$^{\pm.002}$ & 0.612$^{\pm.002}$ & 0.722$^{\pm.002}$ & 3.734$^{\pm.028}$ & 3.703$^{\pm.008}$ & 8.973$^{\pm.071}$ & 0.368$^{\pm.018}$ \\
    TM2T~\cite{guo2022tm2t} $_{\text{ECCV'22}}$ & \checkmark & - & 0.424$^{\pm.003}$ & 0.618$^{\pm.003}$ & 0.729$^{\pm.002}$ & 1.501$^{\pm.017}$ & 3.467$^{\pm.011}$ & 8.589$^{\pm.076}$ & 2.424$^{\pm.093}$ \\
    Guo et al.\cite{humanml3d} $_{\text{CVPR'22}}$ & \checkmark & - & 0.455$^{\pm.003}$ & 0.636$^{\pm.003}$ & 0.736$^{\pm.002}$ & 1.087$^{\pm.021}$ & 3.347$^{\pm.008}$ & 9.175$^{\pm.083}$ & 2.219$^{\pm.074}$ \\
    MotionDiffuse~\cite{zhang2024motiondiffuse} $_{\text{TPAMI'24}}$ & -\checkmark & - & 0.491$^{\pm.001}$ & 0.681$^{\pm.001}$ & 0.782$^{\pm.001}$ & 0.630$^{\pm.001}$ & 3.113$^{\pm.001}$ & 9.410$^{\pm.049}$ & 1.553$^{\pm.042}$ \\
    Fg-T2M~\cite{wang2023fgt2m}  $_{\text{ICCV'23}}$ & \checkmark & - & 0.492$^{\pm.002}$ & 0.683$^{\pm.003}$ & 0.783$^{\pm.002}$ & 0.243$^{\pm.019}$ & 3.109$^{\pm.007}$ & 9.278$^{\pm.072}$ & 1.614$^{\pm.049}$ \\
    FineMoGen~\cite{wang2023fgt2m}  $_{\text{NeurIPS'23}}$ & \checkmark & - & 0.504$^{\pm.002}$ & 0.690$^{\pm.002}$ & 0.784$^{\pm.002}$ & 0.151$^{\pm.008}$ & 2.998$^{\pm.008}$ & 9.263$^{\pm.094}$ & 2.696$^{\pm.079}$ \\

    \midrule[0.5pt]

    MDM~\cite{mdm} $_{\text{arXiv'22}}$ $\dag$ & \checkmark & - & 0.323$^{\pm.006}$ & 0.498$^{\pm.007}$ & 0.606$^{\pm.008}$ & 3.137$^{\pm.183}$ & 4.373$^{\pm.043}$ & 9.525$^{\pm.086}$ & 2.614$^{\pm.102}$ \\

    \textbf{(T\&DT)-MDM} (BPMSD) & \checkmark & \checkmark & 0.445$^{\pm.007}$ & 0.640$^{\pm.009}$ & 0.745$^{\pm.008}$ & 0.756$^{\pm.081}$ & 3.412$^{\pm.030}$ & 9.640$^{\pm.095}$ & 2.495$^{\pm.053}$ \\
    
    \textbf{(T\&DT)-MDM} (BPMP) & \checkmark & \checkmark & 0.460$^{\pm.005}$ & 0.655$^{\pm.005}$ & 0.759$^{\pm.005}$ & 0.488$^{\pm.046}$ & 3.276$^{\pm.021}$ & 9.869$^{\pm.108}$ & 2.340$^{\pm.054}$ \\

    \midrule[0.5pt]
    
    T2M-GPT~\cite{t2mgpt} $_{\text{CVPR'23}}$ $\dag$ & \checkmark & - & 0.499$^{\pm.003}$ & 0.688$^{\pm.003}$ & 0.781$^{\pm.002}$ & 0.123$^{\pm.005}$ & 3.076$^{\pm.009}$ & 9.747$^{\pm.093}$ & 1.890$^{\pm.085}$ \\

    \textbf{(T\&DT)2M-GPT} (BPMSD) & \checkmark & \checkmark & 0.510$^{\pm.002}$ & 0.695$^{\pm.002}$ & 0.789$^{\pm.002}$ & 0.091$^{\pm.004}$ & 3.002$^{\pm.008}$ & 9.592$^{\pm.079}$ & 1.594$^{\pm.075}$ \\
    
    \textbf{(T\&DT)2M-GPT} (BPMP) & \checkmark & \checkmark & 0.506$^{\pm.002}$ & 0.694$^{\pm.002}$ & 0.788$^{\pm.002}$ & 0.100$^{\pm.005}$ & 3.023$^{\pm.010}$ & 9.602$^{\pm.057}$ & 1.615$^{\pm.016}$  \\
    
    \midrule[0.5pt]

    MoMask~\cite{guo2024momask} $_{\text{CVPR'24}}$ $\dag$ & \checkmark & - & 0.466$^{\pm.003}$ & 0.655$^{\pm.003}$ & 0.753$^{\pm.002}$ & 0.249$^{\pm.012}$ & 3.359$^{\pm.008}$ & 9.676$^{\pm.083}$ & 1.371$^{\pm.048}$ \\

    \textbf{(T\&DT)-MoMask} (BPMSD) & \checkmark & \checkmark & 0.519$^{\pm.002}$ & 0.715$^{\pm.002}$ & 0.811$^{\pm.001}$ & 0.088$^{\pm.003}$ & 2.946$^{\pm.005}$ & 9.702$^{\pm.075}$ & 1.271$^{\pm.030}$ \\
    
    \textbf{(T\&DT)-MoMask} (BPMP) & \checkmark & \checkmark & 0.520$^{\pm.003}$ & 0.717$^{\pm.002}$ & 0.813$^{\pm.002}$ & 0.055$^{\pm.002}$ & 2.935$^{\pm.009}$ & 9.679$^{\pm.085}$ & 1.281$^{\pm.051}$ \\

    \bottomrule[1pt]
    
\end{tabular}
\setlength{\abovecaptionskip}{0.3em} 
\caption{
    \textbf{Benchmark of FineMotion \& Comparisons with HumanML3D.}
    We conduct all evaluations 20 times, reporting the average with a 95\% confidence interval, except for MModality, which is run 5 times. 
    `$\rightarrow$' means results are better if the metric is closer to the real motions.
    For methods marked with $\dag$, we re-implement them using the same text encoder (T5) as ours to ensure fair comparisons. 
    All our variants exhibit performance improvements, with (T\&DT)-MDM showing a notable +15.3\% increase in Top-3 retrieval accuracy.
} \label{table: humanml3d_t2m}
\vspace{-2em}
\end{center}
\end{table*}

\vspace{0.5em}
\noindent\textbf{Human Annotations.} 
Similarly, we train (T\&DT)2M-GPT on both the T2M and (T\&DT)2M tasks, using human annotations as DT (`\textbf{DT$^\text{Human}$}') when available and automatically generated ones otherwise.
From Tab.~\ref{table: ablation}.(3) and (5),
human annotations offer more precise guidance, and further enhance performance, achieving an FID of 0.091 and an R-Top3 of 0.789 for BPMSD, and an FID of 0.100 and an R-Top3 of 0.788 for BPMP.
In the following experiments, we adopt the training setting from Tab.~\ref{table: ablation}.(3) and (5), and use (T\&DT$^\text{Human}$)2M as the default test configuration.

\subsection{Impact on Text-driven Motion Generation} 

\textbf{Benchmarking FineMotion.}
From Tab.~\ref{table: humanml3d_t2m}, including our fine-grained texts improves all variants.
(T\&DT)-MoMask achieves the best overall performance but the lowest MModality score, indicating reduced motion diversity.
(T\&DT)2M-GPT performs competitively while preserving high MModality, demonstrating our dataset’s potential to enhance GPT-based methods.
(T\&DT)-MDM attains the highest MModality but the lowest R-Precision, suggesting it generates noisy and jittery motions.

\vspace{0.5em}
\noindent\textbf{Comparison with HumanML3D.}
To validate the significance of our dataset, we conduct a comparative analysis between FineMotion and HumanML3D (Coarse Text Only, \textit{i.e.}, `T') in Tab.~\ref{table: humanml3d_t2m}.
We re-implement MDM~\cite{mdm}, T2M-GPT~\cite{t2mgpt}, and MoMask~\cite{guo2024momask} on HumanML3D, replacing their text encoders with T5-Base~\cite{t5} for fair comparisons.
Overall, all variants trained on FineMotion consistently outperform those using HumanML3D’s coarse descriptions (\dag), achieving better FID and R-Precision.
Notably, MDM improves Top-3 retrieval accuracy by +15.3\%, while (T\&DT)-MoMask achieves the best FID and R-Precision across both datasets.
For the suboptimal Diversity and MModality of our variants, we attribute this to the fine-grained descriptions, which constrain motion variation. 
These results underscore the versatility of FineMotion and its potential for zero-shot fine-grained motion editing.


\subsection{Zero-shot Fine-Grained Motion Editing}
\label{sec:edit_results}

Existing motion editing works rely on coarse captions that lack detail, limiting control over body part movements, timing, and duration.
To address this, we use (T\&DT)2M-GPT as an example and train it with pairs of coarse captions, temporally augmented motions, and corresponding detailed texts.
The augmented data enhances the model's ability to align motions with BPM snippet descriptions. 
Thus, (T\&DT)2M-GPT model can precisely control body part actions at specific time intervals based on detailed texts.

\begin{figure}[!t]
\begin{center}
\includegraphics[width=1.0\linewidth]{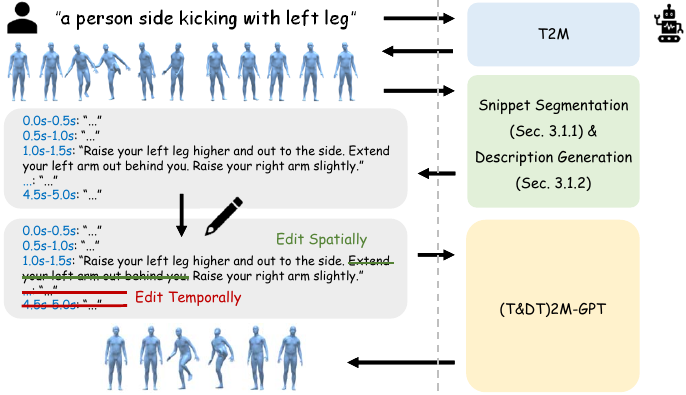}
\end{center}
\vspace{-1.2em}
\caption{
    \textbf{Pipeline for zero-shot fine-grained motion editing.} 
    To edit human motion with fine granularity, users first provide a coarse textual description of the desired motion. An initial motion is generated using any text-to-motion (T2M) model. This motion is then processed through the dataset construction pipeline to extract its BPM snippet descriptions. Users refine these descriptions with detailed editing instructions. Finally, the baseline model generates the fine-grained edited motion by adhering to both the modified BPM snippets and the original coarse caption.
}
\label{fig:editing_pipeline}
\vspace{-1em}
\end{figure}

Based on this, we achieve the effect of fine-grained motion editing by the zero-shot pipeline in illustrated Fig.~\ref{fig:editing_pipeline}.
It begins with users providing a text-to-motion model with a coarse description to synthesize an initial motion.
Detailed descriptions are then obtained following the dataset construction pipeline, allowing users to refine them with fine-grained editing requirements. 
Finally, (T\&DT)2M-GPT generates a new motion sequence from scratch based on the modified description, producing the final edited result. 
Notably, the re-generation process may introduce unintended changes beyond the specified regions.

\begin{figure}[!b]
\vspace{-0.5em}
\begin{center}
\includegraphics[width=1.0\linewidth]{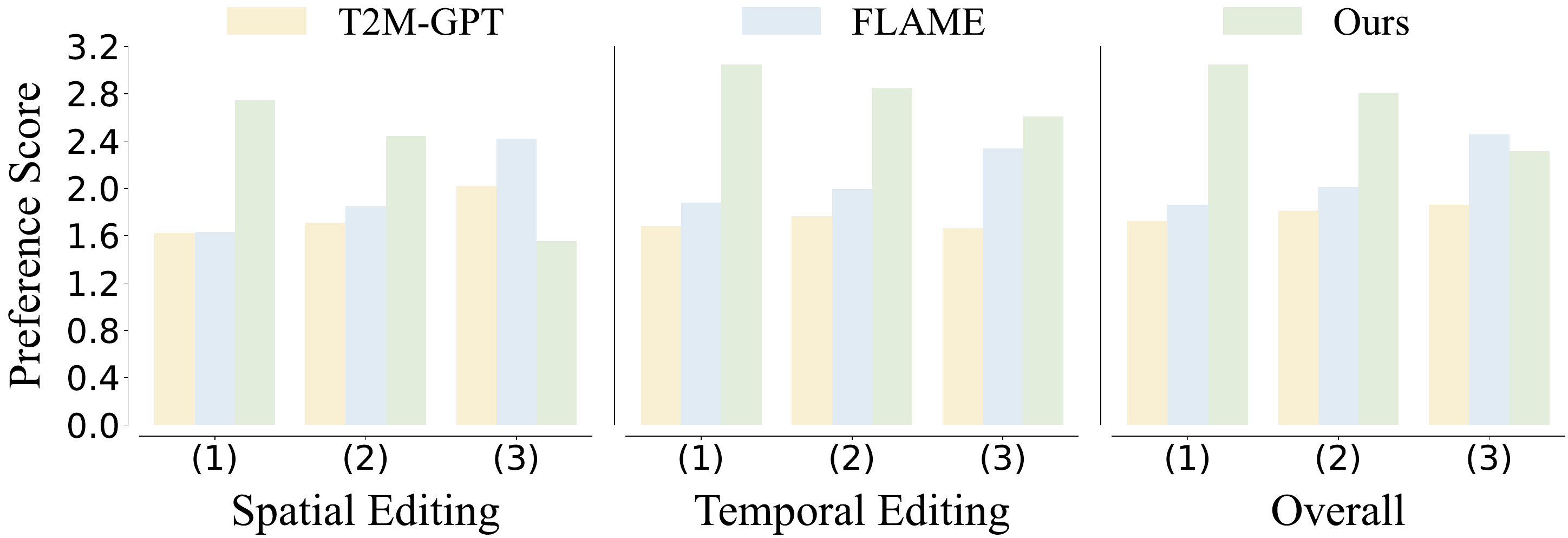}
\setlength{\abovecaptionskip}{-0.5em} 
\end{center}
\vspace{-1.5em}
   \caption{
        \textbf{The statistical results of the user study.}
        The \textit{left} figure displays the average preference score for the spatial editing results (3 cases) of three models, with a score of 3 for the best model, 2 for the second, and 1 for the last.
        The \textit{middle} one shows the score for the temporal editing results (6 cases).
        The \textit{right} one summarizes the results for all 9 cases.
        Each case is evaluated from three perspectives:
        (1) whether the edited motion meets the editing requirements,
        (2) the naturalness of the edited motion,
        and (3) the similarity between the edited motion and the original one.
   }
    \label{fig:user_study_statistics}
\end{figure}

Since quantitative metrics for evaluating editing results are unavailable, we follow \cite{kim2023flame} and \cite{mdm} by conducting a user study with 30 randomly selected participants. 
Each user ranked 9 cases across 3 perspectives, totaling 27 questions. 
We compare our editing pipeline against T2M-GPT~\cite{t2mgpt}, a generative-based motion generation model, and FLAME~\cite{kim2023flame}, a diffusion-based text-driven motion editing approach. 
As these models lack fine-grained textual training data, their editing results are generated using coarse captions describing the post-edited motions.

\begin{figure}[!t]
\begin{center}
\includegraphics[width=1.0\linewidth]{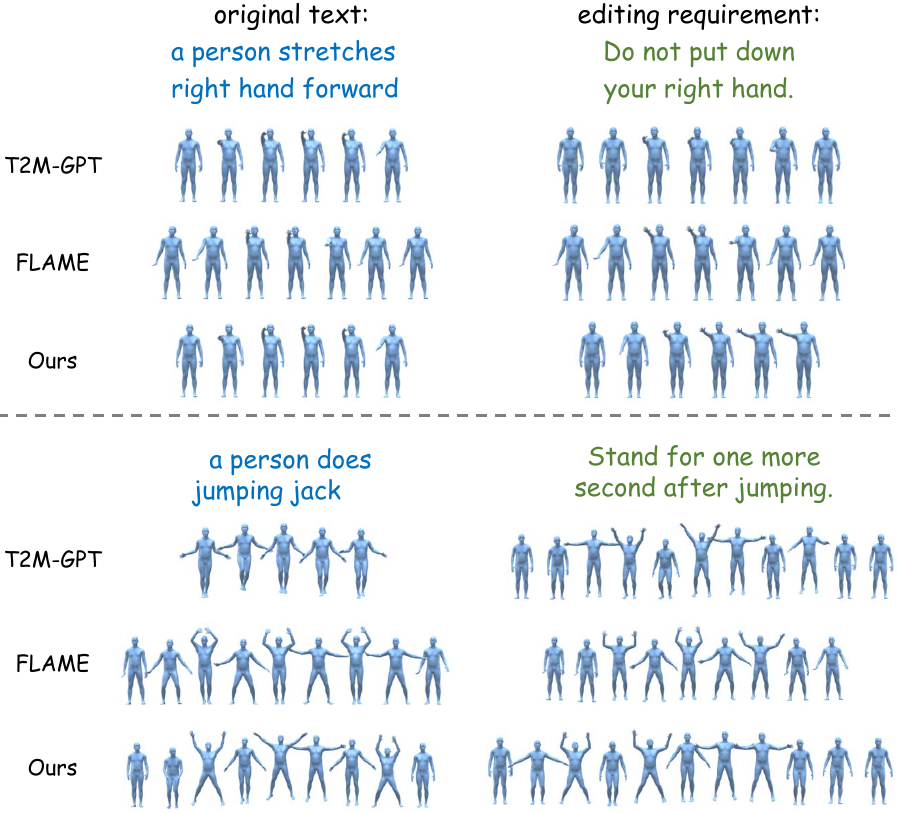}
\end{center}
\vspace{-1.5em}
   \caption{
        Examples of editing results where our pipeline achieves the highest preference score in terms of \textit{meeting the editing requirements}.
        \textbf{Top:} Spatial Editing.
        \textbf{Bottom:} Temporal Editing.
   }
    \label{fig:user_study_examples}
\vspace{-1em}
\end{figure}

As shown in Fig.~\ref{fig:user_study_statistics}, users clearly preferred the edited results from our pipeline, as the other two methods often failed to meet the editing requirements. 
Our edited motions were also rated as the most natural. 
In terms of similarity to the original motions, our pipeline ranked second, as the other methods frequently produced identical outputs (\textit{i.e.}, higher similarity) due to their inability to perform edits effectively.
Notably, our pipeline excels in temporal editing, particularly in adjusting motion length, a task the other methods struggled with.
This advantage stems from the temporally augmented data used in training (T\&DT)2M-GPT, making temporal editing easier than spatial editing.
Fig.~\ref{fig:user_study_examples} presents examples where our pipeline's edits were clearly favored by users compared to other methods. 
The user study is presented in the appendix.

\section{Conclusion}
\label{sec:conclusion}
This paper introduces FineMotion, a comprehensive dataset featuring human motion sequences paired with BPM snippet descriptions and paragraphs.
We also develop an automated annotation pipeline to enable efficient dataset scaling. 
To validate the dataset's significance, we adapt three classical text-to-motion methods and benchmark them using our detailed textual annotations. 
Experimental results demonstrate that our detailed text improves motion generation performance, paving the way for zero-shot fine-grained motion editing.
A user study confirms the high quality of our editing results. 
We anticipate that this exploratory work will shed light on future research in developing effective fine-grained motion understanding systems.



\section*{Acknowledgements}
This work was supported in part by National Key R\&D Program of China (No. 2024YFF0618400), National Natural Science Foundation of China under Grant 82261138629, Shenzhen Municipal Science and Technology Innovation Council under Grant JCYJ20220531101412030, Guangdong Provincial Key Laboratory under Grant 2023B1212060076, and the Ningbo Municipal Bureau of Science and Technology under Grant
2023Z138, 2023Z237, 2024Z110 and 2024Z124.

{
    \small
    \bibliographystyle{ieeenat_fullname}
    \bibliography{main}

\begin{thebibliography}{34}
\providecommand{\natexlab}[1]{#1}
\providecommand{\url}[1]{\texttt{#1}}
\expandafter\ifx\csname urlstyle\endcsname\relax
  \providecommand{\doi}[1]{doi: #1}\else
  \providecommand{\doi}{doi: \begingroup \urlstyle{rm}\Url}\fi

\bibitem[Athanasiou et~al.()Athanasiou, Petrovich, Black, and Varol]{athanasiousinc}
Nikos Athanasiou, Mathis Petrovich, Michael~J Black, and G{\"u}l Varol.
\newblock Sinc: Spatial composition of 3d human motions for simultaneous action generation supplementary material.

\bibitem[Delmas et~al.(2022)Delmas, Weinzaepfel, Lucas, Moreno-Noguer, and Rogez]{delmas2022posescript}
Ginger Delmas, Philippe Weinzaepfel, Thomas Lucas, Francesc Moreno-Noguer, and Gr{\'e}gory Rogez.
\newblock Posescript: 3d human poses from natural language.
\newblock In \emph{European Conference on Computer Vision}, pages 346--362. Springer, 2022.

\bibitem[Delmas et~al.(2023)Delmas, Weinzaepfel, Moreno-Noguer, and Rogez]{delmas2023posefix}
Ginger Delmas, Philippe Weinzaepfel, Francesc Moreno-Noguer, and Gr{\'e}gory Rogez.
\newblock Posefix: Correcting 3d human poses with natural language.
\newblock In \emph{Proceedings of the IEEE/CVF International Conference on Computer Vision}, pages 15018--15028, 2023.

\bibitem[Grammarly()]{grammarly}
Grammarly.
\newblock \url{https://www.grammarly.com}.

\bibitem[Guo et~al.(2020)Guo, Zuo, Wang, Zou, Sun, Deng, Gong, and Cheng]{humanact12}
Chuan Guo, Xinxin Zuo, Sen Wang, Shihao Zou, Qingyao Sun, Annan Deng, Minglun Gong, and Li Cheng.
\newblock Action2motion: Conditioned generation of 3d human motions.
\newblock In \emph{Proceedings of the 28th ACM International Conference on Multimedia}, pages 2021--2029, 2020.

\bibitem[Guo et~al.(2022{\natexlab{a}})Guo, Zou, Zuo, Wang, Ji, Li, and Cheng]{humanml3d}
Chuan Guo, Shihao Zou, Xinxin Zuo, Sen Wang, Wei Ji, Xingyu Li, and Li Cheng.
\newblock Generating diverse and natural 3d human motions from text.
\newblock In \emph{Proceedings of the IEEE/CVF Conference on Computer Vision and Pattern Recognition}, pages 5152--5161, 2022{\natexlab{a}}.

\bibitem[Guo et~al.(2022{\natexlab{b}})Guo, Zuo, Wang, and Cheng]{guo2022tm2t}
Chuan Guo, Xinxin Zuo, Sen Wang, and Li Cheng.
\newblock Tm2t: Stochastic and tokenized modeling for the reciprocal generation of 3d human motions and texts.
\newblock In \emph{European Conference on Computer Vision}, pages 580--597. Springer, 2022{\natexlab{b}}.

\bibitem[Guo et~al.(2024)Guo, Mu, Javed, Wang, and Cheng]{guo2024momask}
Chuan Guo, Yuxuan Mu, Muhammad~Gohar Javed, Sen Wang, and Li Cheng.
\newblock Momask: Generative masked modeling of 3d human motions.
\newblock In \emph{Proceedings of the IEEE/CVF Conference on Computer Vision and Pattern Recognition}, pages 1900--1910, 2024.

\bibitem[He et~al.(2023)He, Huang, Zhan, Wen, and Shan]{he2023semanticboost}
Xin He, Shaoli Huang, Xiaohang Zhan, Chao Wen, and Ying Shan.
\newblock Semanticboost: Elevating motion generation with augmented textual cues.
\newblock \emph{arXiv preprint arXiv:2310.20323}, 2023.

\bibitem[Jiang et~al.(2024)Jiang, Chen, Liu, Yu, Yu, and Chen]{openmotionlab_motiongpt}
Biao Jiang, Xin Chen, Wen Liu, Jingyi Yu, Gang Yu, and Tao Chen.
\newblock Motiongpt: Human motion as a foreign language.
\newblock \emph{Advances in Neural Information Processing Systems}, 36, 2024.

\bibitem[Kalakonda et~al.(2023)Kalakonda, Maheshwari, and Sarvadevabhatla]{kalakonda2023actiongpt}
Sai~Shashank Kalakonda, Shubh Maheshwari, and Ravi~Kiran Sarvadevabhatla.
\newblock Action-gpt: Leveraging large-scale language models for improved and generalized action generation.
\newblock In \emph{2023 IEEE International Conference on Multimedia and Expo (ICME)}, pages 31--36. IEEE, 2023.

\bibitem[Kim et~al.(2023)Kim, Kim, and Choi]{kim2023flame}
Jihoon Kim, Jiseob Kim, and Sungjoon Choi.
\newblock Flame: Free-form language-based motion synthesis \& editing.
\newblock In \emph{Proceedings of the AAAI Conference on Artificial Intelligence}, pages 8255--8263, 2023.

\bibitem[Li et~al.(2022)Li, Zhao, Zhelun, and Sheng]{li2022danceformer}
Buyu Li, Yongchi Zhao, Shi Zhelun, and Lu Sheng.
\newblock Danceformer: Music conditioned 3d dance generation with parametric motion transformer.
\newblock In \emph{Proceedings of the AAAI Conference on Artificial Intelligence}, pages 1272--1279, 2022.

\bibitem[Li et~al.(2024)Li, Dai, Zhang, Li, Yang, Guo, and Li]{li2024exploring}
Ronghui Li, Yuqin Dai, Yachao Zhang, Jun Li, Jian Yang, Jie Guo, and Xiu Li.
\newblock Exploring multi-modal control in music-driven dance generation.
\newblock \emph{arXiv preprint arXiv:2401.01382}, 2024.

\bibitem[Loper et~al.(2023)Loper, Mahmood, Romero, Pons-Moll, and Black]{loper2023smpl}
Matthew Loper, Naureen Mahmood, Javier Romero, Gerard Pons-Moll, and Michael~J Black.
\newblock Smpl: A skinned multi-person linear model.
\newblock In \emph{Seminal Graphics Papers: Pushing the Boundaries, Volume 2}, pages 851--866. 2023.

\bibitem[Mahmood et~al.(2019)Mahmood, Ghorbani, Troje, Pons-Moll, and Black]{mahmood2019amass}
Naureen Mahmood, Nima Ghorbani, Nikolaus~F Troje, Gerard Pons-Moll, and Michael~J Black.
\newblock Amass: Archive of motion capture as surface shapes.
\newblock In \emph{Proceedings of the IEEE/CVF international conference on computer vision}, pages 5442--5451, 2019.

\bibitem[Petrovich et~al.(2021)Petrovich, Black, and Varol]{actor}
Mathis Petrovich, Michael~J Black, and G{\"u}l Varol.
\newblock Action-conditioned 3d human motion synthesis with transformer vae.
\newblock In \emph{Proceedings of the IEEE/CVF International Conference on Computer Vision}, pages 10985--10995, 2021.

\bibitem[Petrovich et~al.(2022)Petrovich, Black, and Varol]{petrovich2022temos}
Mathis Petrovich, Michael~J Black, and G{\"u}l Varol.
\newblock Temos: Generating diverse human motions from textual descriptions.
\newblock In \emph{European Conference on Computer Vision}, pages 480--497. Springer, 2022.

\bibitem[Plappert et~al.(2016)Plappert, Mandery, and Asfour]{kitml}
Matthias Plappert, Christian Mandery, and Tamim Asfour.
\newblock The kit motion-language dataset.
\newblock \emph{Big data}, 4\penalty0 (4):\penalty0 236--252, 2016.

\bibitem[Raab et~al.(2023)Raab, Leibovitch, Li, Aberman, Sorkine-Hornung, and Cohen-Or]{raab2023modi}
Sigal Raab, Inbal Leibovitch, Peizhuo Li, Kfir Aberman, Olga Sorkine-Hornung, and Daniel Cohen-Or.
\newblock Modi: Unconditional motion synthesis from diverse data.
\newblock In \emph{Proceedings of the IEEE/CVF Conference on Computer Vision and Pattern Recognition}, pages 13873--13883, 2023.

\bibitem[Radford et~al.(2021)Radford, Kim, Hallacy, Ramesh, Goh, Agarwal, Sastry, Askell, Mishkin, Clark, et~al.]{radford2021clip}
Alec Radford, Jong~Wook Kim, Chris Hallacy, Aditya Ramesh, Gabriel Goh, Sandhini Agarwal, Girish Sastry, Amanda Askell, Pamela Mishkin, Jack Clark, et~al.
\newblock Learning transferable visual models from natural language supervision.
\newblock In \emph{International conference on machine learning}, pages 8748--8763. PMLR, 2021.

\bibitem[Raffel et~al.(2020)Raffel, Shazeer, Roberts, Lee, Narang, Matena, Zhou, Li, and Liu]{t5}
Colin Raffel, Noam Shazeer, Adam Roberts, Katherine Lee, Sharan Narang, Michael Matena, Yanqi Zhou, Wei Li, and Peter~J Liu.
\newblock Exploring the limits of transfer learning with a unified text-to-text transformer.
\newblock \emph{Journal of machine learning research}, 21\penalty0 (140):\penalty0 1--67, 2020.

\bibitem[Team et~al.(2023)Team, Anil, Borgeaud, Wu, Alayrac, Yu, Soricut, Schalkwyk, Dai, Hauth, et~al.]{team2023gemini}
Gemini Team, Rohan Anil, Sebastian Borgeaud, Yonghui Wu, Jean-Baptiste Alayrac, Jiahui Yu, Radu Soricut, Johan Schalkwyk, Andrew~M Dai, Anja Hauth, et~al.
\newblock Gemini: a family of highly capable multimodal models.
\newblock \emph{arXiv preprint arXiv:2312.11805}, 2023.

\bibitem[Tevet et~al.(2022{\natexlab{a}})Tevet, Gordon, Hertz, Bermano, and Cohen-Or]{tevet2022motionclip}
Guy Tevet, Brian Gordon, Amir Hertz, Amit~H Bermano, and Daniel Cohen-Or.
\newblock Motionclip: Exposing human motion generation to clip space.
\newblock In \emph{European Conference on Computer Vision}, pages 358--374. Springer, 2022{\natexlab{a}}.

\bibitem[Tevet et~al.(2022{\natexlab{b}})Tevet, Raab, Gordon, Shafir, Cohen-Or, and Bermano]{mdm}
Guy Tevet, Sigal Raab, Brian Gordon, Yonatan Shafir, Daniel Cohen-Or, and Amit~H Bermano.
\newblock Human motion diffusion model.
\newblock \emph{arXiv preprint arXiv:2209.14916}, 2022{\natexlab{b}}.

\bibitem[Tseng et~al.(2023)Tseng, Castellon, and Liu]{tseng2023edge}
Jonathan Tseng, Rodrigo Castellon, and Karen Liu.
\newblock Edge: Editable dance generation from music.
\newblock In \emph{Proceedings of the IEEE/CVF Conference on Computer Vision and Pattern Recognition}, pages 448--458, 2023.

\bibitem[Van Den~Oord et~al.(2017)Van Den~Oord, Vinyals, et~al.]{van2017vqvae}
Aaron Van Den~Oord, Oriol Vinyals, et~al.
\newblock Neural discrete representation learning.
\newblock \emph{Advances in neural information processing systems}, 30, 2017.

\bibitem[Wang et~al.(2023)Wang, Leng, Li, Wu, and Liang]{wang2023fgt2m}
Yin Wang, Zhiying Leng, Frederick~WB Li, Shun-Cheng Wu, and Xiaohui Liang.
\newblock Fg-t2m: Fine-grained text-driven human motion generation via diffusion model.
\newblock In \emph{Proceedings of the IEEE/CVF International Conference on Computer Vision}, pages 22035--22044, 2023.

\bibitem[Yazdian et~al.(2023)Yazdian, Liu, Cheng, and Lim]{yazdian2023motionscript}
Payam~Jome Yazdian, Eric Liu, Li Cheng, and Angelica Lim.
\newblock Motionscript: Natural language descriptions for expressive 3d human motions.
\newblock \emph{arXiv preprint arXiv:2312.12634}, 2023.

\bibitem[Zhang et~al.(2023)Zhang, Zhang, Cun, Zhang, Zhao, Lu, Shen, and Shan]{t2mgpt}
Jianrong Zhang, Yangsong Zhang, Xiaodong Cun, Yong Zhang, Hongwei Zhao, Hongtao Lu, Xi Shen, and Ying Shan.
\newblock Generating human motion from textual descriptions with discrete representations.
\newblock In \emph{Proceedings of the IEEE/CVF Conference on Computer Vision and Pattern Recognition}, pages 14730--14740, 2023.

\bibitem[Zhang et~al.(2024{\natexlab{a}})Zhang, Cai, Pan, Hong, Guo, Yang, and Liu]{zhang2024motiondiffuse}
Mingyuan Zhang, Zhongang Cai, Liang Pan, Fangzhou Hong, Xinying Guo, Lei Yang, and Ziwei Liu.
\newblock Motiondiffuse: Text-driven human motion generation with diffusion model.
\newblock \emph{IEEE Transactions on Pattern Analysis and Machine Intelligence}, 2024{\natexlab{a}}.

\bibitem[Zhang et~al.(2024{\natexlab{b}})Zhang, Li, Cai, Ren, Yang, and Liu]{zhang2024finemogen}
Mingyuan Zhang, Huirong Li, Zhongang Cai, Jiawei Ren, Lei Yang, and Ziwei Liu.
\newblock Finemogen: Fine-grained spatio-temporal motion generation and editing.
\newblock \emph{Advances in Neural Information Processing Systems}, 36, 2024{\natexlab{b}}.

\bibitem[Zhang et~al.(2019)Zhang, Kishore, Wu, Weinberger, and Artzi]{zhang2019bertscore}
Tianyi Zhang, Varsha Kishore, Felix Wu, Kilian~Q Weinberger, and Yoav Artzi.
\newblock Bertscore: Evaluating text generation with bert.
\newblock \emph{arXiv preprint arXiv:1904.09675}, 2019.

\bibitem[Zhang et~al.(2024{\natexlab{c}})Zhang, Huang, Liu, Tang, Lu, Chen, Bai, Chu, Yu, and Ouyang]{zhang2024motiongpt}
Yaqi Zhang, Di Huang, Bin Liu, Shixiang Tang, Yan Lu, Lu Chen, Lei Bai, Qi Chu, Nenghai Yu, and Wanli Ouyang.
\newblock Motiongpt: Finetuned llms are general-purpose motion generators.
\newblock In \emph{Proceedings of the AAAI Conference on Artificial Intelligence}, pages 7368--7376, 2024{\natexlab{c}}.

\end{thebibliography}
}

\clearpage
\setcounter{page}{1}
\maketitlesupplementary

\section{License}
\label{sec:license}
The license for human motion sequences in this dataset follows the term specified at \href{https://github.com/EricGuo5513/HumanML3D/blob/main/LICENSE}{HumanML3D} and \href{https://amass.is.tue.mpg.de/license.html}{AMASS}.
The textual descriptions in our FineMotion dataset are under the CC BY 4.0 International license. 
For detailed license information, please refer to
\url{https://creativecommons.org/licenses/by/4.0/legalcode}

\section{Discussion on Selecting of optimal $T_s$}
To determine the optimal snippet duration $T_s$, we propose two guiding principles to help researchers tailor this value to their own datasets.
As shown in Fig.~\ref{fig:statistics_snippet}, we randomly sample 1,000 snippets with varying durations from all motion sequences in our dataset.
Then, we calculate the cosine similarity between the PoseScript~\cite{delmas2022posescript} semantic features of the start and end poses for each snippet. 
A higher cosine similarity indicates that the start and end poses are more similar, suggesting that the motion progresses slowly; conversely, a lower similarity indicates faster progression.

Our results show that the motions in our dataset generally progress slowly, prompting the selection of a larger interval to avoid redundancy.
Here, we also display the statistical results of a rapidly changing motion (1,000 random samples of start and end points) in Fig.~\ref{fig:statistics_snippet}.
The results indicate that the similarity of the pose semantic features first decreases and then increases as the temporal interval grows.
From this, we derive the first principle for selecting the optimal value of $T_s$: \textbf{Choose the value of $T_s$ that minimizes the similarity between the start and the end poses.}
Meanwhile, PoseFix~\cite{delmas2023posefix} suggests that larger time differences between two poses allow for a wide range of plausible in-between motions.
Therefore, the second principle is that \textbf{the value of $T_s$ should not exceed 0.5s}, which is the maximum time difference for pose pair selection specified by PoseFix~\cite{delmas2023posefix}.
Following these principles, we set $T_s$ to 0.5s.
Notably, any remaining segment of a motion sequence shorter than $T_s$ is also treated as an individual snippet.

\section{Data Format Examples}
\label{sec:data_format_example}

The data format example for all the detailed human body part snippet descriptions (BPMSDs) in a whole human motion sequence is shown below:

\begin{tcolorbox}[boxrule=0pt, colframe=white, sharp corners, left=1mm, right=1mm, top=1mm, bottom=1mm]

\begin{scriptsize}

    \{

    \qquad"000314": \qquad\qquad\textcolor{lightgray}{\# name of motion sequence}
    \par
    \qquad[
    \par
    \qquad\qquad"",\qquad\qquad\textcolor{lightgray}{\# 0.0s-0.5s' BPMSD}
    \par
    \qquad\qquad"Bend your elbows and raise your hands up to your head.",
    \par
    \qquad\qquad"",\qquad\qquad\textcolor{lightgray}{\# 1.0s-1.5s' BPMSD}
    \par
    \qquad\qquad"",\qquad\qquad\textcolor{lightgray}{\# 1.5s-2.0s' BPMSD}
    \par
    \qquad\qquad"Turn your upper body to the right slightly.",
    \par
    \qquad\qquad"",\qquad\qquad\textcolor{lightgray}{\# 2.5s-3.0s' BPMSD}
    \par
    \qquad\qquad"Straighten your elbows and lower your hands to your thighs.",
    \par
    \begin{adjustwidth}{4em}{0pt}
    "Straighten your elbows completely and move your hands back to your sides.",
    \end{adjustwidth}
    \par
    \qquad],
    \par
    \}

\end{scriptsize}
\end{tcolorbox}

The data format example for three different detailed human body part paragraphs (BPMPs) for the same human motion sequence is shown below:

\begin{tcolorbox}[boxrule=0pt, colframe=white, sharp corners, left=1mm, right=1mm, top=1mm, bottom=1mm]
\begin{scriptsize}

    \{

    \qquad"000314": \qquad\qquad\textcolor{lightgray}{\# name of motion sequence}
    \par
    \qquad[
    \par
    \begin{adjustwidth}{4em}{0pt}
    "Initially, the person bends his elbows and raises his hands to his head. Then, he slightly turns his upper body to the right. Afterward, he straightens his elbows and lowers his hands to his thighs. Finally, he straightens his elbows completely and moves his hands back to his sides.",
    \end{adjustwidth}
    \par
    \vspace{0.1em}
    \begin{adjustwidth}{4em}{0pt}
    "First, the person bends the elbows and raises his hands above his head. Then, he slightly rotates his upper body to the right. Subsequently, he straightens the elbows and lowers his hands to rest on his thighs. Finally, he fully extends his elbows and returns his hands to their positions at his sides.",
    \end{adjustwidth}
    \par
    \vspace{0.1em}
    \begin{adjustwidth}{4em}{0pt}
    "The person begins by bending the elbows and raising the hands toward the head. Subsequently, he slightly twists his upper body to the right. Afterward, he extends the elbows and lowers the hands toward the thighs, then fully straightening the elbows and moving the hands back to the sides."
    \end{adjustwidth}
    \par
    \qquad],
    \par
    \}
    
\end{scriptsize}
    
\end{tcolorbox}

\clearpage
\newpage

\onecolumn

\section{More Dataset Examples}
\begin{figure*}[!h]
\begin{center}
\includegraphics[width=1.0\linewidth]{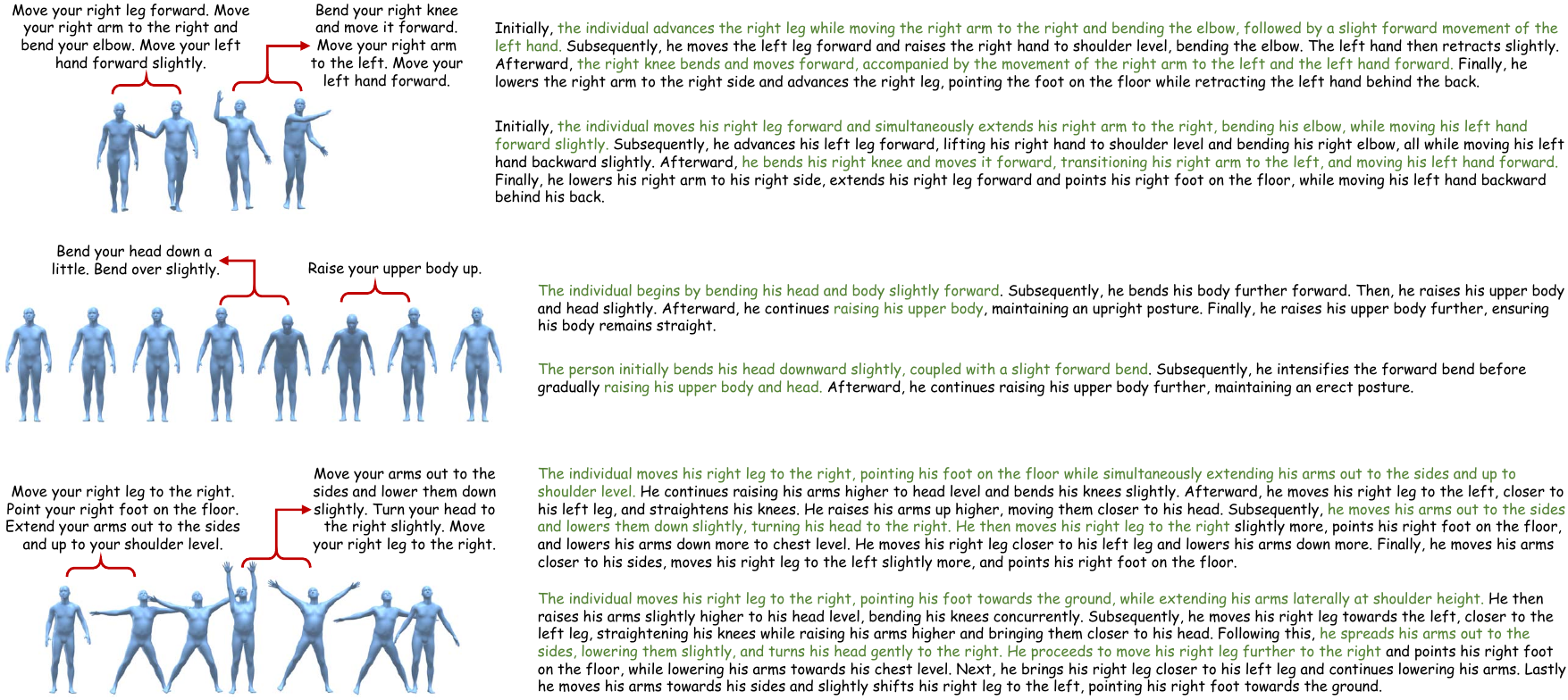}
\end{center}
   \caption{
        More examples of human-annotated body part movement snippet descriptions (\textit{left}) and paragraphs (\textit{right}).
        The colored text in paragraphs links to corresponding snippet descriptions.
   }
    \label{fig:dataset_examples_SM_human}
\end{figure*}

\begin{figure*}[!h]
\begin{center}
\includegraphics[width=1.0\linewidth]{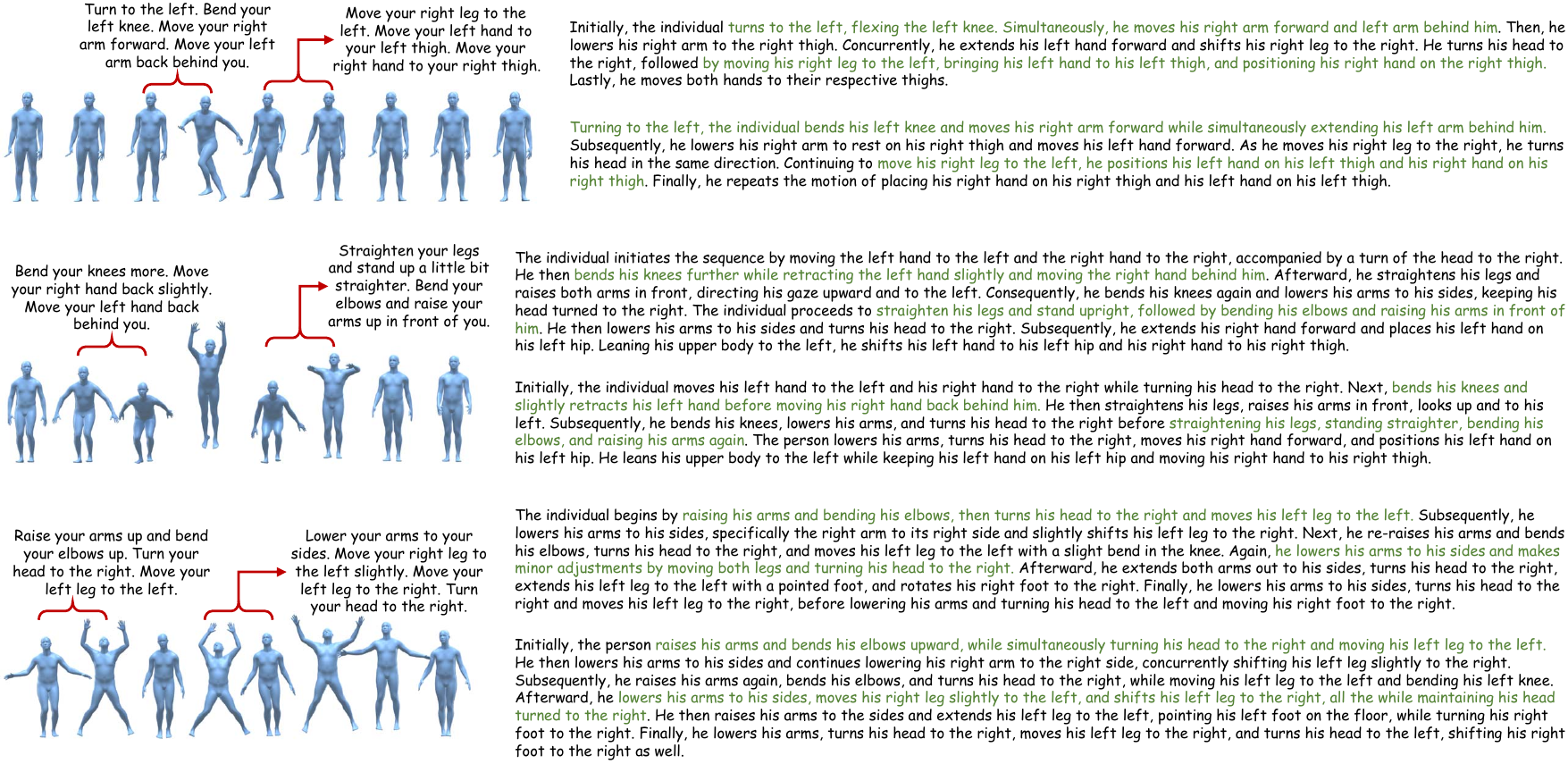}
\end{center}
   \caption{
        More examples of automatically generated body part movement snippet descriptions (\textit{left}) and paragraphs (\textit{right}).
        The colored text in paragraphs links to corresponding snippet descriptions.
   }
    \label{fig:dataset_examples_SM_auto}
\end{figure*}

We display more examples of body part movement descriptions for motion snippet (\textit{i.e.}, BPMSD) and for whole motion sequence (\textit{i.e.}, BPMP) of our FineMotion dataset in Fig.~ \ref{fig:dataset_examples_SM_human} and \ref{fig:dataset_examples_SM_auto}.

\setlength{\floatsep}{5pt} 

\twocolumn

\clearpage
\newpage

\section{Baseline Model Details}
\label{sec:model_details}

This section outlines the network architecture and the implementation of three variants of motion generation methods, including MDM~\cite{mdm}, T2M-GPT~\cite{t2mgpt}, and MoMask~\cite{guo2024momask} on our dataset, and denoted them as (T\&DT)-MDM, (T\&DT)2M-GPT, and (T\&DT)-MoMask, respectively.

\begin{itemize}
    \item \textbf{(T\&DT)-MDM} builds from MDM~\cite{mdm}. 
    It employs a classifier-free, diffusion-based approach for human motion generation using a transformer-based architecture. 
    Unlike standard diffusion models, it directly predicts the sample at each diffusion step. 
    Specifically, the transformer-encoder predicts the final clean motion based on a condition (\textit{i.e.}, a CLIP-based textual embedding), a noising timestep, and random noise.
    To accommodate detailed textual descriptions, which often contain over ten times the number of tokens compared to coarse captions,
    we replace the CLIP~\cite{radford2021clip} text encoder with the T5-Base~\cite{t5} encoder, which uses relative attention for flexible input lengths.
    We then perform mean pooling along the sequence length dimension of the T5-Base encoder output to obtain a single text embedding for each text.
    Now, the model's condition turns out to be the concatenated text embeddings of both the coarse caption and the detailed description.

    \item \textbf{(T\&DT)2M-GPT} is derived from T2M-GPT~\cite{t2mgpt} and comprises a Motion VQ-VAE and a GPT model.
    Motion VQ-VAE learns a mapping between raw motion sequences and discrete token sequences, while the GPT model generates motion tokens conditioned on text embeddings.
    Likewise, we modified the condition of the GPT model into the concatenated T5 text embeddings of the coarse caption and the detailed text.

    \item \textbf{(T\&DT)-MoMask} is based on MoMask~\cite{guo2024momask}, featuring a Motion Residual VQ-VAE, a Masked Transformer, and a Residual Transformer.
    Concretely, the Residual VQ-VAE uses a hierarchical quantization scheme to discretize motions into multiple layers of motion tokens.
    The Masked Transformer predicts masked motion tokens from the text input, 
    while the Residual Transformer progressively predicts next-layer tokens based on the results from the current layer.
    The textual embedding is modified similarly to the previous two networks.
\end{itemize}

All three baseline models are adapted to include our long, detailed body part movement descriptions for the motion sequences.
Here, we hold (T\&DT)2M-GPT as the example to elaborate on the differences from the original T2M-GPT model.
The modifications applied to the other two baseline models, (T\&DT)-MDM and (T\&DT)-MoMask, follow a similar approach.

\begin{figure}[!h]
\begin{center}
\includegraphics[width=1.0\linewidth]{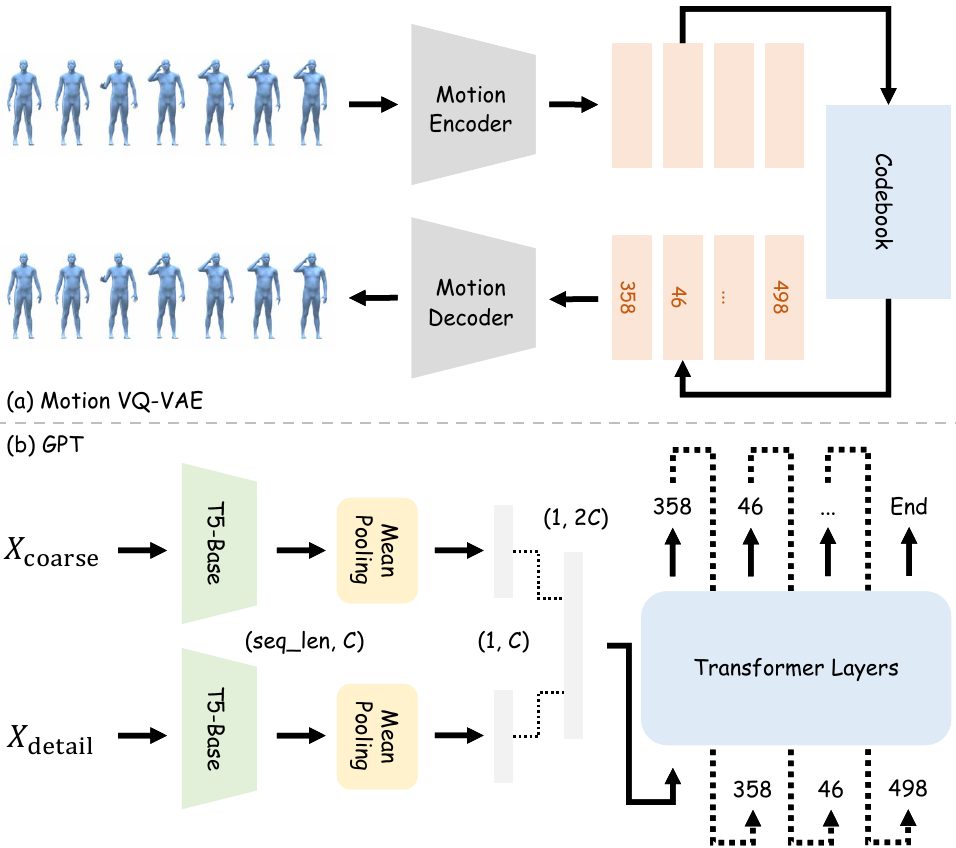}
\end{center}
   \caption{
        \textbf{Overview of the baseline network, (T\&DT)2M-GPT.}
        It generates motions that strictly follow the fine-grained description $X_\text{detail}$ and the coarse-grained caption $X_\text{coarse}$.
        It consists of a motion VQ-VAE for discretizing the motion into tokens and a GPT for generating motion tokens.
   }
    \label{fig:baseline_model}
\end{figure}

(T\&DT)2M-GPT mainly contains two parts: Motion VQ-VAE for motion discretization and GPT for generating motion tokens from the coarse caption and detailed text.

\vspace{0.5em}
\noindent\textbf{Motion VQ-VAE.}
We follow~\cite{t2mgpt} to represent motions in discrete tokens, and vice versa.
Specifically, it contains an encoder $E$, a decoder $D$, and a learnable codebook $B=\{b_k\}^K_{k=1}$, where $K$ is the size of the codebook. 
Given a $T$-frame motion sequence $M = [m_1, m_2, \dots, m_T]$ with $m_t \in \mathbb{R}^d$, the encoder $E$ maps it into a sequence of latent features $Z = E(M)$ with $Z = [z_1, z_2, \dots, z_{\lfloor T/l \rfloor}]$ and $z_i \in \mathbb{R}^{d_c}$, where $l$ represents the temporal downsampling rate of the encoder $E$.
Then, these latent features are transformed into a sequence of motion codes $C = [c_1, c_2, \dots, c_{\lfloor T/l \rfloor}]$, where $c_i$ is the index of the most similar element to $z_i$ in $B$.
With a sequence of motion codes $C$, 
we first project $C$ back to their corresponding codebook elements $\widetilde{Z} = [\widetilde{z}_1, \widetilde{z}_2, \dots, \widetilde{z}_{\lfloor T/l \rfloor}]$ with $\widetilde{z}_i = b_{c_i}$. Then, the decoder $D$ reconstructs $\widetilde{Z}$ into a motion sequence 
$\widetilde{M} = D(\widetilde{Z}) = [\widetilde{m}_1, \widetilde{m}_2, \dots, \widetilde{m}_T]$. 
The motion VQ-VAE is optimized by the standard optimization goal~\cite{van2017vqvae} that requires the decoded motion $\widetilde{M}$ to be as similar as the input motion $M$.
The exponential moving average (EMA) and codebook reset (Code Reset) are employed to stabilize the training process.
With a learned motion VQ-VAE, a motion sequence can be easily mapped into discrete motion tokens by the encoder $E$ and the codebook $B$.
On the other hand, the output of our (T\&DT)2M-GPT model, \textit{i.e.}, motion tokens, can be recovered into motion sequences by the decoder $D$ and the codebook $B$.

\vspace{0.5em}
\noindent\textbf{GPT for generating motion tokens.}
First, we extract the text embeddings of the coarse caption $t_\text{coarse}$ and the edited detailed motion script $\hat{t}_\text{detail}$.
Since the number of tokens of our detailed human body part descriptions is usually more than ten times that of the coarse captions, 
we use the frozen encoder from T5-Base~\cite{t5} to extract the textual embeddings, considering that its relative attention mechanism allows input with any sequence length.
We then perform a mean pooling operation in the \textit{seq\_len} dimension of the output from the T5-Base encoder to obtain a single text embedding for each text.
\begin{equation}
    t_\text{coarse} = \text{Mean}(\text{T5Encoder}(X_\text{coarse})) \in \mathbb{R}^{768},
\end{equation}
\begin{equation}
    \hat{t}_\text{detail} = \text{Mean}(\text{T5Encoder}(\hat{X}_\text{detail})) \in \mathbb{R}^{768}.
\end{equation}

Next, the two text embeddings are utilized as the conditions of the GPT model to autoregressively generate motion tokens.
The GPT model is composed of a stack of transformer layers. 
Besides, casual self-attention is applied to ensure the calculation of the current tokens does not consider the information of the future motion tokens.
Since this fine-grained motion generation task can be considered as the next motion token prediction task, which is based on the given coarse textual embedding, the motion script textual embedding, and previous motion tokens, 
the GPT model is optimized by the cross-entropy loss between the predicted motion tokens and ground-truth ones.
\begin{equation}
    L = - \sum_{i=1}^{\lfloor T/l \rfloor} log(P(c_i \mid t_\text{coarse}, \hat{t}_\text{detail}, c_{<i}, \theta_\text{GPT})).
\end{equation}

After sufficient training, the GPT model can generate appropriate motion tokens that can be further decoded into motions by the decoder in Motion VQ-VAE.

\section{More Implementation Details}
\label{sec:implementation_hyperparameters}

The architecture and training hyperparameters of our baseline models strictly follow those in the original paper~\cite{t2mgpt, guo2024momask, mdm}. 
Notably, since we replace the text encoder with that of T5, the dimension of output text embedding turns to 768 rather than that of the CLIP text encoder, 512.
Therefore, the input of the fully connected layer that projects the CLIP text embedding to the input of the GPT also needs to be changed from 512 to 768.
The code is based on PyTorch.
The experiments were conducted on the A100-80G GPU, but only about 16G GPU memory was used.
Due to replacing the text encoder with a larger model~\cite{t5} and using it to process longer textual descriptions, the training time for (T\&DT)2M-GPT increases to 154 hours, compared to the 78 hours reported in~\cite{t2mgpt} for T2M-GPT. 
However, the training time can be reduced to the original 78 hours if all text embeddings are pre-extracted and stored before the training begins.

\section{More Discussion on Motion Generation with Fine-grained Texts Only}
We did not evaluate this setting because it will lead to \textbf{ambiguity} in motion generation. 
Fine-grained text captures detailed body part movements and timing, while coarse text supplements global motion semantics, both crucial for precise motion generation.
For instance, motions with coarse text `\textit{a person is standing still}' and `\textit{a person is sitting}' share the same fine-grained text (\textless Motionless\textgreater, \textit{i.e.}, no body part movements).
The model cannot distinguish such cases without coarse text, degrading motion generation performance.
Given the issue above, we do not train our models using (fine-grained text, motion) pairs. 
Evaluating such a setting without proper training would lead to unfair or unreliable results.

\section{More Discussion on Table 2}
One may notice that when (T\&DT)2M-GPT—\textit{i.e.}, Rows (2)-(5) in Table 2—generates motions using only coarse descriptions (Test Set: T2M), it shows a slight performance drop, compared to our implementation of T2M-GPT trained solely on the T2M task, Row (1). 
The slight drop in T2M-GPT variants likely stems from their high sensitivity to the shared training budget, as multi-task training with (T\&DT) halves the T2M updates compared to the baseline.
Additional evaluations on MDM and MoMask variants show that including (T\&DT)2M during training actually improves motion generation when only coarse text is available, as shown below.

\begin{table}[!h]
\begin{center}
\resizebox{\columnwidth}{!}{
\begin{tabular}{ccccccccc}
    \toprule[1pt]

    \multicolumn{2}{c}{Train Task} & Test Task & \multicolumn{2}{c}{MDM} & \multicolumn{2}{c}{T2M-GPT} & \multicolumn{2}{c}{MoMask} \\

    \cmidrule(r){1-2}\cmidrule(r){3-3}\cmidrule(r){4-5}\cmidrule(r){6-7}\cmidrule(r){8-9}
    
    T2M & (T\&DT)2M & T2M & R-Top3 $\uparrow$ & FID $\downarrow$ & R-Top3 $\uparrow$ & FID $\downarrow$ & R-Top3 $\uparrow$ & FID $\downarrow$ \\
    
    \midrule[0.5pt]

    \checkmark & - & \checkmark & 0.606$^{\pm.008}$ & 3.137$^{\pm.183}$ & 0.781$^{\pm.003}$ & 0.123$^{\pm.005}$ & 0.753$^{\pm.002}$ & 0.249$^{\pm.012}$ \\

    \checkmark & (our BPMSD) & \checkmark & 0.746$^{\pm.007}$ & 0.760$^{\pm.064}$ & 0.781$^{\pm.002}$ & 0.154$^{\pm.007}$ & 0.827$^{\pm.002}$ & 0.120$^{\pm.004}$ \\

    \checkmark & (our BPMP) & \checkmark & 0.759$^{\pm.006}$ & 0.436$^{\pm.043}$ & 0.781$^{\pm.002}$ & 0.155$^{\pm.006}$ & 0.818$^{\pm.002}$ & 0.130$^{\pm.005}$ \\

    \bottomrule[1pt]
\end{tabular}
}
\setlength{\abovecaptionskip}{5pt} 
\caption{
    Generation performance of all our variants on the T2M test set, \textit{i.e.}, motion generation conditioned on coarse descriptions only.
}
\end{center}
\end{table}

\section{Ablation Study on Baseline Model Design}
Here, we conduct an ablation study on different strategies for encoding coarse and detailed texts.
Specifically, we denote the strategy of connecting the coarse text (T) and detailed text (DT) into a single text and feeding it into the text encoder as `TDT'. 
Meanwhile, `T\&DT' refers to encoding T and DT separately and then concatenating their resulting embeddings.
Results below show that the `TDT' strategy leads to poorer performance, likely because the model is overwhelmed by the dense information and struggles to capture the global motion semantics. These findings highlight that our baseline designs are carefully considered, rather than naïve implementations.

\begin{table}[!h]
\begin{center}
\resizebox{\columnwidth}{!}{
\begin{tabular}{ccccccc}
    \toprule[1pt]
     \multirow{2}{*}{Method} & \multicolumn{3}{c}{R-Precision $\uparrow$} & \multirow{2}{*}{FID $\downarrow$}  & \multirow{2}{*}{MM-Dist $\downarrow$}  & \multirow{2}{*}{Diversity $\rightarrow$}    \\

    \cmidrule(r){2-4}

     & Top-1 & Top-2 & Top-3  \\

    \midrule[1pt]

    TDT-MoMask (BPMSD) & 0.212$^{\pm.002}$ & 0.341$^{\pm.002}$ & 0.434$^{\pm.002}$ & 8.328$^{\pm.056}$ & 5.877$^{\pm.009}$ & 8.899$^{\pm.069}$ \\

    (T\&DT)-MoMask (BPMSD) & 0.519$^{\pm.002}$ & 0.715$^{\pm.002}$ & 0.811$^{\pm.001}$ & 0.088$^{\pm.003}$ & 2.946$^{\pm.005}$ & 9.702$^{\pm.075}$ \\

    \midrule[1pt]

    TDT-MoMask (BPMP) & 0.358$^{\pm.003}$ & 0.528$^{\pm.002}$ & 0.628$^{\pm.002}$ & 0.285$^{\pm.006}$ & 4.145$^{\pm.008}$ & 9.626$^{\pm.093}$  \\

    (T\&DT)-MoMask (BPMP) & 0.520$^{\pm.003}$ & 0.717$^{\pm.002}$ & 0.813$^{\pm.002}$ & 0.055$^{\pm.002}$ & 2.935$^{\pm.009}$ & 9.679$^{\pm.085}$ \\

    \bottomrule[1pt]
\end{tabular}
}
\setlength{\abovecaptionskip}{5pt} 
\caption{
    Ablation study on different strategies for encoding coarse and detailed texts.
}
\end{center}
\end{table}

\section{Metrics and Results for Temporal Alignment}
Currently, there is no metric that directly evaluates the precision of temporal alignment between detailed texts and generated motion sequences.
Given that our detailed texts are strictly aligned with ground-truth motions over time, we reframe this evaluation as measuring the alignment between short clips of generated motions and corresponding ground-truth clips.
High similarity between these clips—even at fine temporal granularity—implies accurate alignment with the detailed texts.

To this end, we introduce FID$_c$, which computes the similarity between generated and ground-truth motions using overlapping 40-frame clips (the minimum evaluation length), with a stride of 10—matching the minimal temporal interval of our detailed texts.
The table below reports FID$_c$ scores across all clips. As shown, our variants (last two rows) achieve significantly lower FID$_c$ scores, demonstrating that our generated motions are better temporally aligned with the detailed texts, compared to motions generated by models trained solely on coarse descriptions.

\begin{table}[!h]
\centering
\footnotesize
\setlength{\tabcolsep}{5pt}
\begin{tabular}{cccccccccc}

    \toprule[1pt]
    
     & MDM & T2M-GPT & MoMask \\ 
     
    \midrule[0.5pt] 
    
    T2M & 3.012$^{\pm.206}$ & 1.423$^{\pm.040}$ & 0.293$^{\pm.011}$ & \\
    (T\&DT)2M (BPMSD) & 1.382$^{\pm.125}$ & 0.398$^{\pm.011}$ & 0.165$^{\pm.004}$ \\
    (T\&DT)2M (BPMP) & 0.426$^{\pm.046}$ & 0.624$^{\pm.015}$ & 0.108$^{\pm.003}$ \\
    
    \bottomrule[1pt]
    
\end{tabular}
\setlength{\abovecaptionskip}{5pt} 
\caption{
    Comparison of temporal alignment, measured by FID$_c$, between baseline text-to-motion models and our fine-grained variants.
}
\end{table}

\section{Limitations and Future Work}
\label{sec:limit_future}
Since we use temporally augmented data to train the text-to-motion models, editing motions along the temporal dimension becomes more straightforward and accurate compared to spatial editing.
Consequently, future work will focus on developing effective methods for spatial human motion editing. 

Additionally, obtaining the detailed body part textual descriptions still requires multiple steps. 
Thus, training an end-to-end model that can directly infer these descriptions from human motion sequences presents a promising research direction.

Moreover, the capabilities of large language models (LLMs) could be leveraged to unify text-to-motion and motion-to-text tasks through textual descriptions of varying granularity, potentially enhancing the effectiveness of both tasks.

\clearpage
\newpage

\onecolumn

\section{User Study}
\label{sec:users_study}

\textbf{Case 1}: \textbf{Add} the body part movements \textbf{Spatially}.

\begin{figure*}[!h]
\vspace{-1em}
\begin{center}
\includegraphics[width=0.7\linewidth]{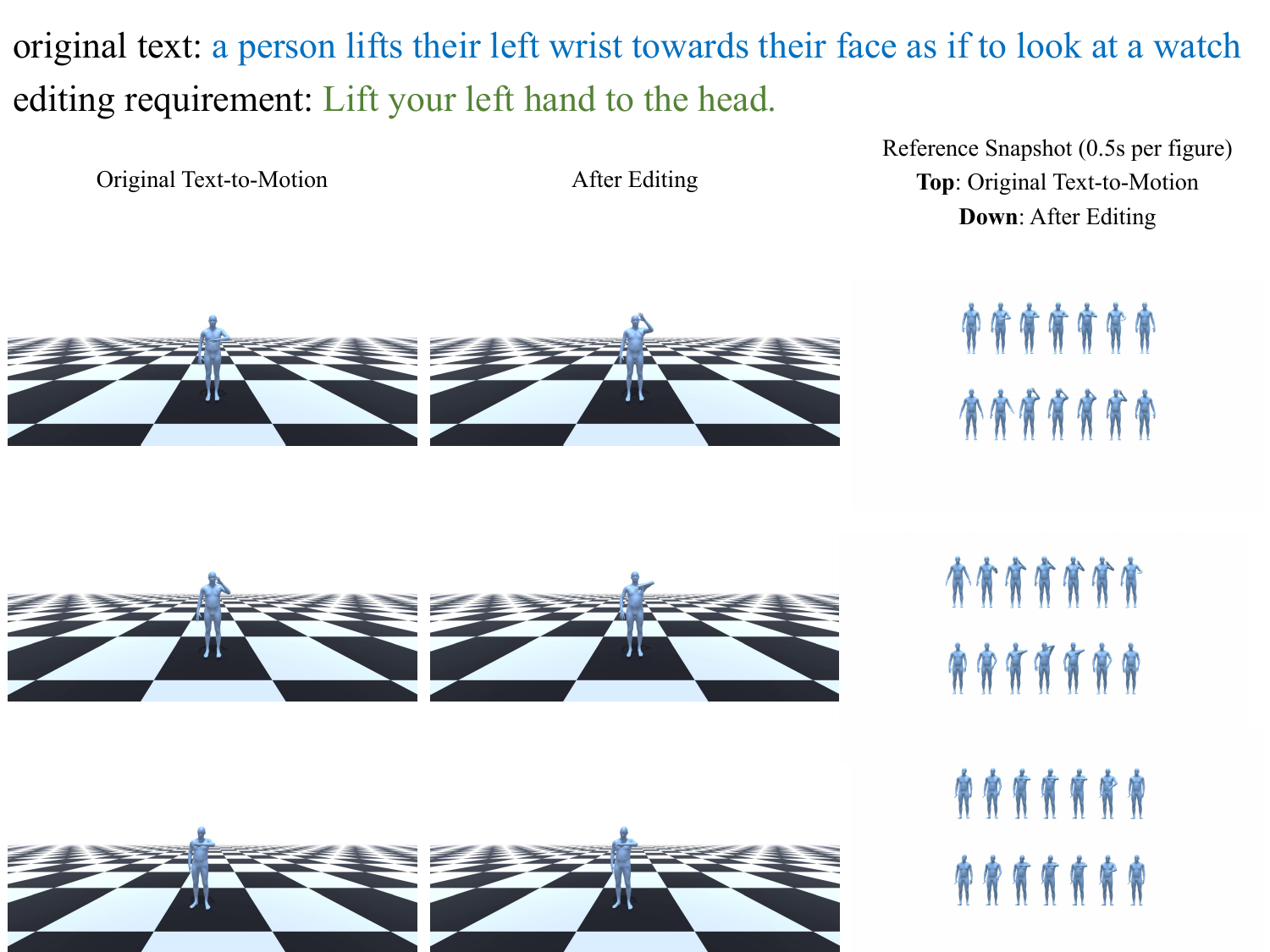}
\end{center}
\end{figure*}

\textcolor{red}{Answer: \quad Row 1: Ours \quad Row 2: T2M-GPT \quad Row 3: FLAME}

\vspace{1em}
\textbf{Case 2}: \textbf{Delete} the body part movements \textbf{Spatially}.

\begin{figure*}[!h]
\vspace{-1em}
\begin{center}
\includegraphics[width=0.7\linewidth]{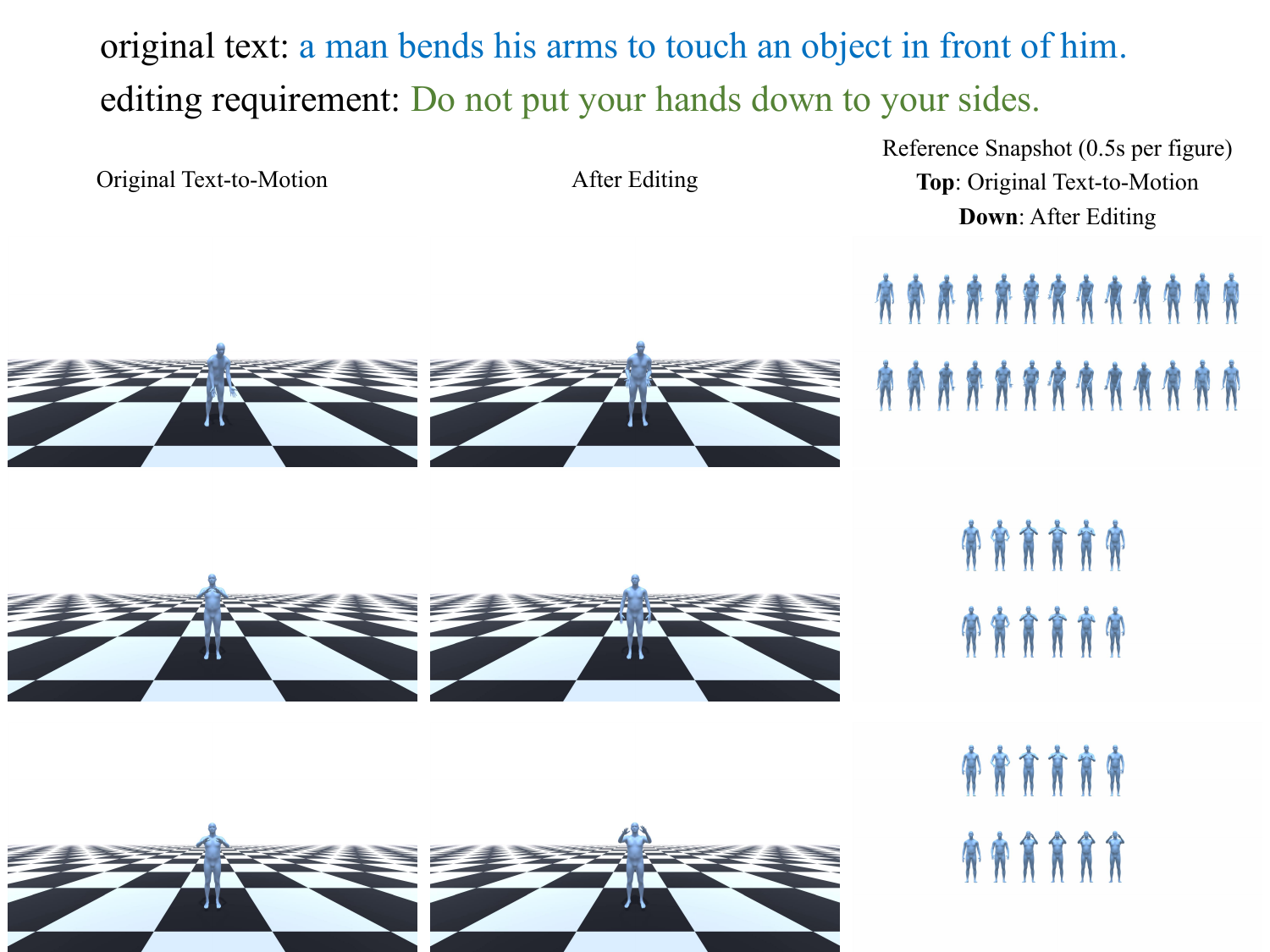}
\end{center}
\end{figure*}

\textcolor{red}{Answer: \quad Row 1: FLAME \quad Row 2: T2M-GPT \quad Row 3: Ours}

\newpage
\textbf{Case 3}: \textbf{Modify} the body part movements \textbf{Spatially}.

\begin{figure*}[!h]
\vspace{-1em}
\begin{center}
\includegraphics[width=0.7\linewidth]{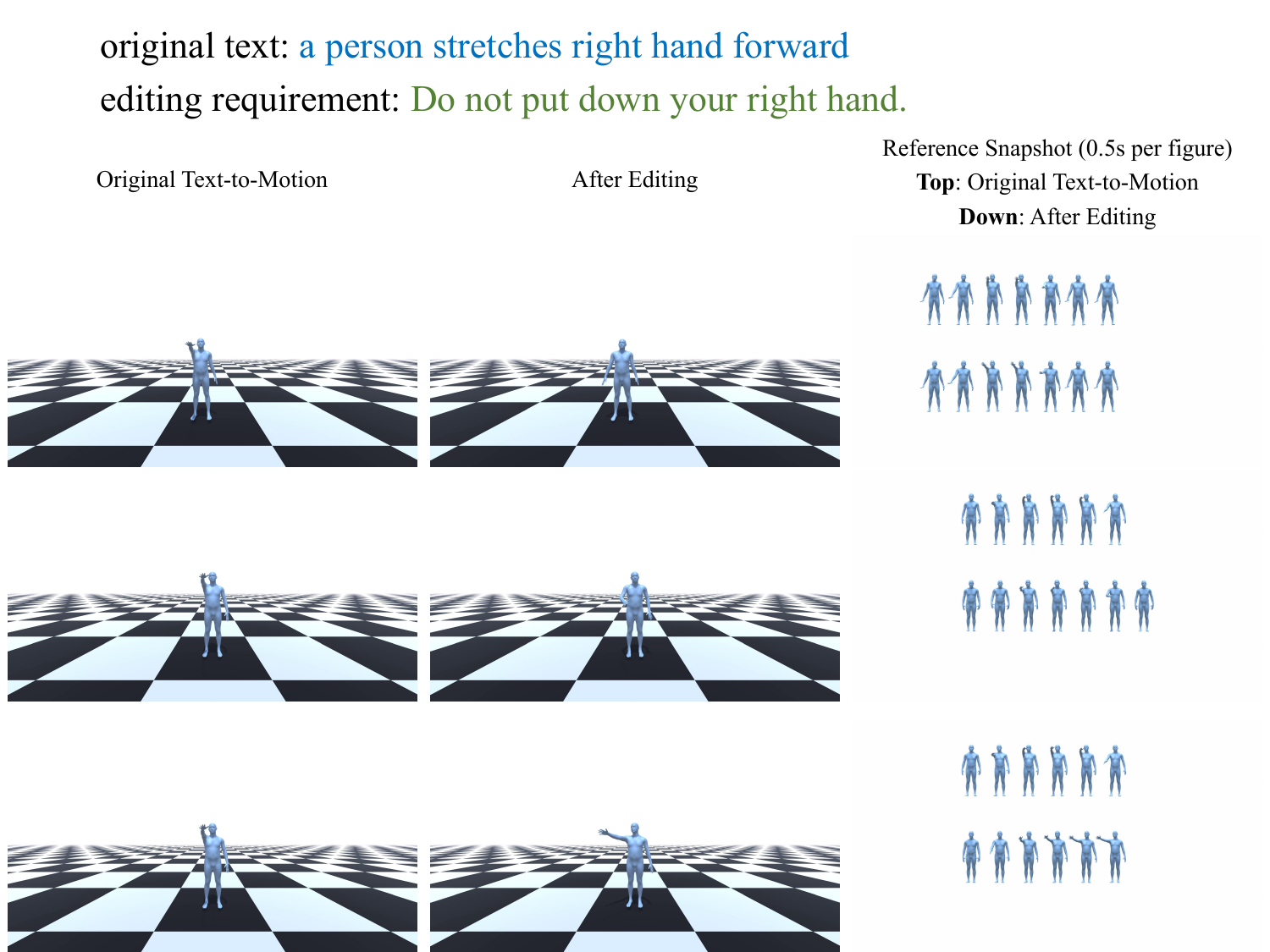}
\end{center}
\end{figure*}

\textcolor{red}{Answer: \quad Row 1: FLAME \quad Row 2: T2M-GPT \quad Row 3: Ours}

\vspace{1em}
\textbf{Case 4}: \textbf{Extend} at the \textbf{start} of the human motion (\textbf{Temporally}).

\begin{figure*}[!h]
\vspace{-1em}
\begin{center}
\includegraphics[width=0.7\linewidth]{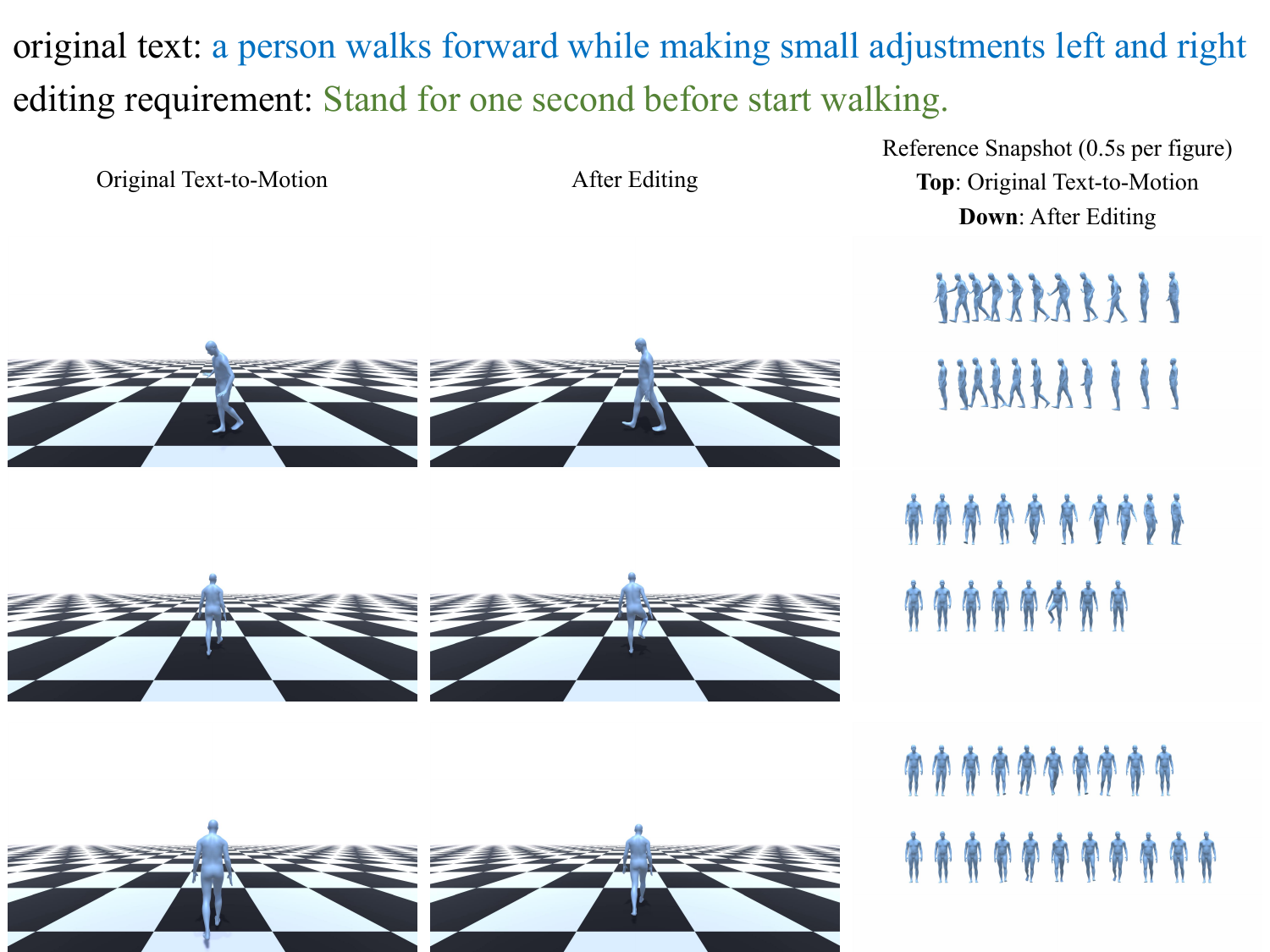}
\end{center}
\end{figure*}

\textcolor{red}{Answer: \quad Row 1: FLAME \quad Row 2: T2M-GPT \quad Row 3: Ours}

\newpage
\textbf{Case 5}: \textbf{Delete} at the \textbf{end} of the human motion (\textbf{Temporally}).

\begin{figure*}[!h]
\vspace{-1em}
\begin{center}
\includegraphics[width=0.7\linewidth]{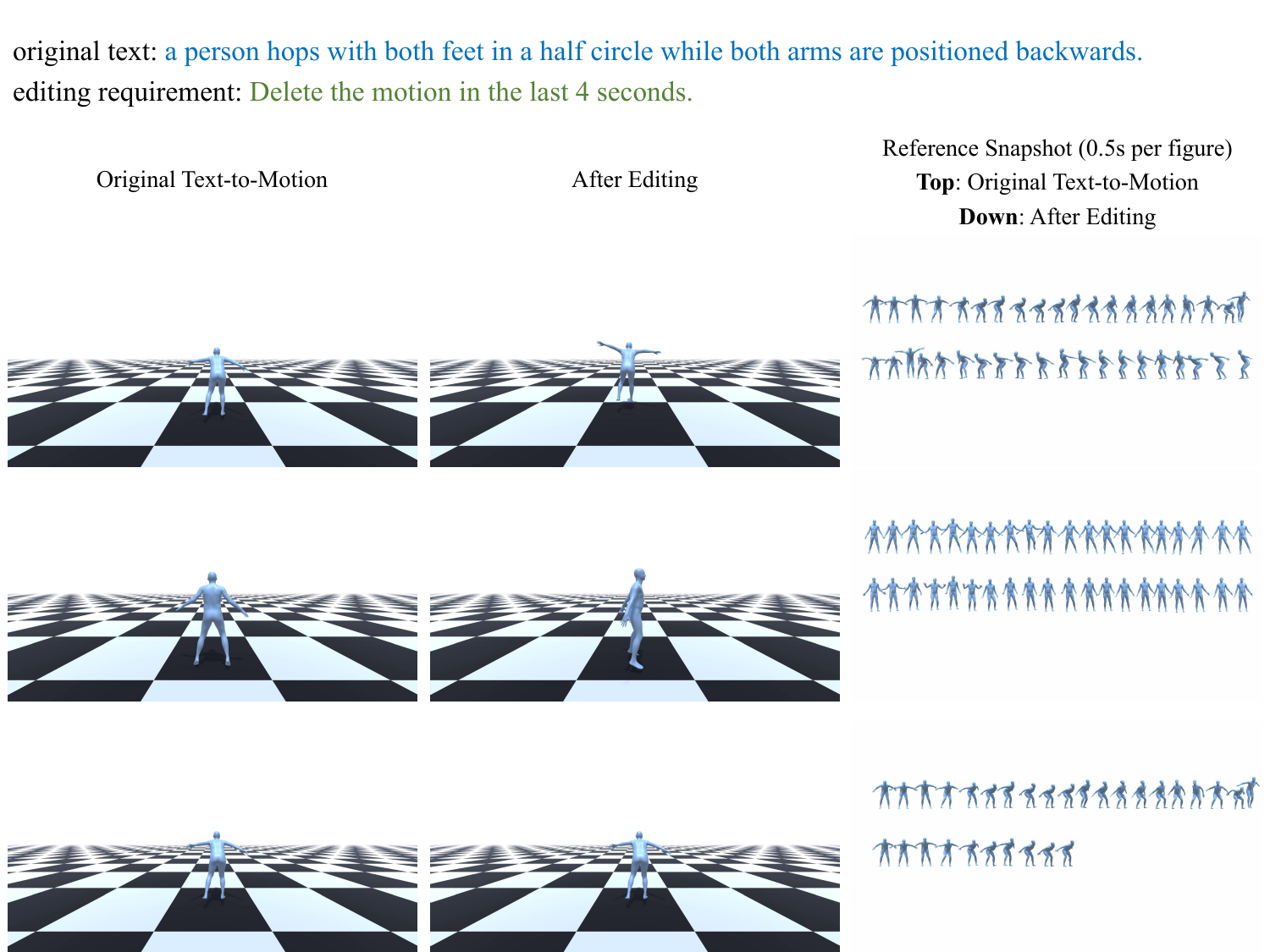}
\end{center}
\end{figure*}

\textcolor{red}{Answer: \quad Row 1: T2M-GPT \quad Row 2: FLAME \quad Row 3: Ours}

\vspace{1em}
\textbf{Case 6}: \textbf{Delete} in the \textbf{middle} of the human motion (\textbf{Temporally}).

\begin{figure*}[!h]
\vspace{-1em}
\begin{center}
\includegraphics[width=0.7\linewidth]{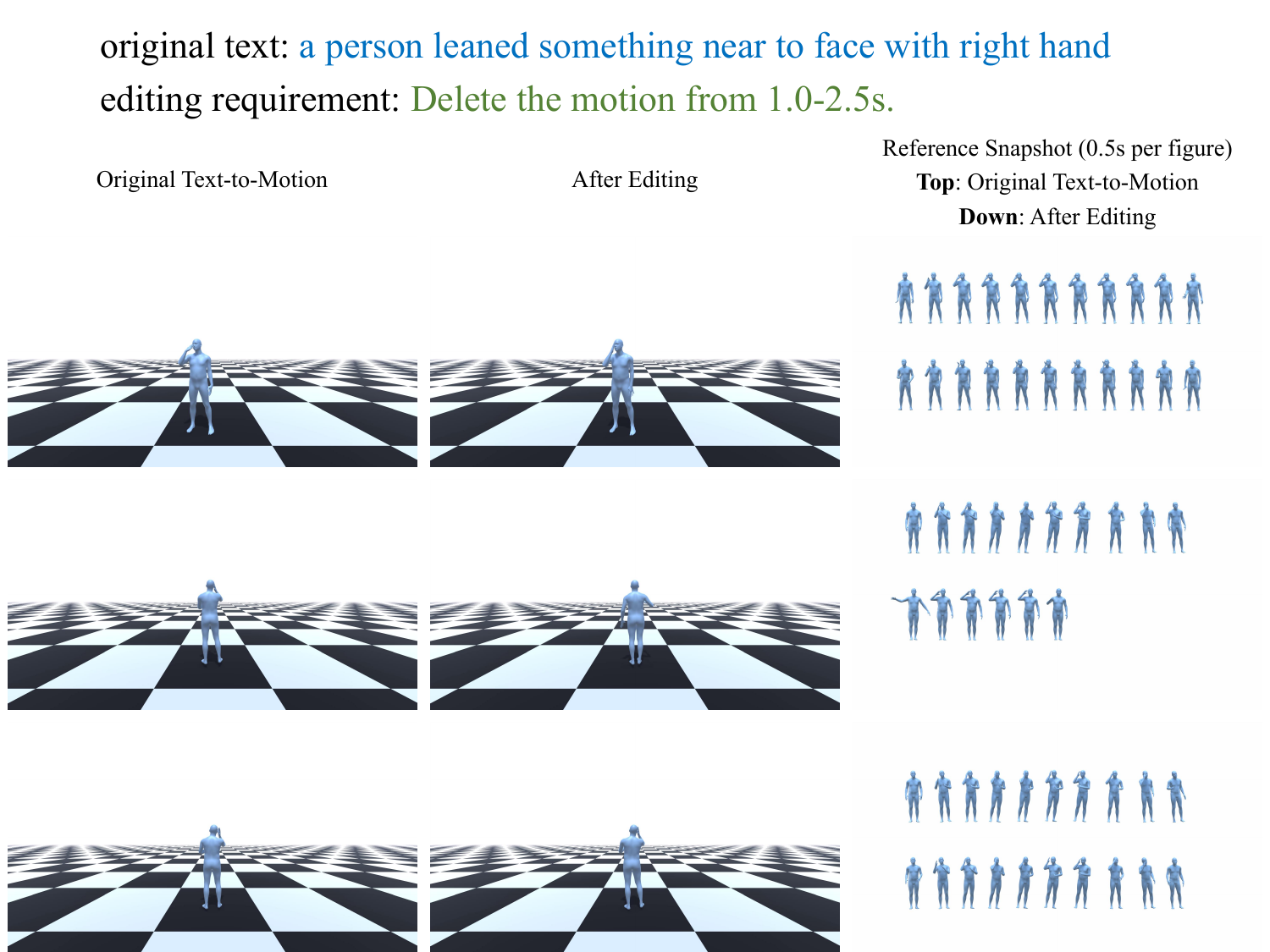}
\end{center}
\end{figure*}

\textcolor{red}{Answer: \quad Row 1: FLAME \quad Row 2: Ours \quad Row 3: T2M-GPT}

\newpage
\textbf{Case 7}: \textbf{Insert} in the middle of the human motion (\textbf{Temporally}).

\begin{figure*}[!h]
\vspace{-1em}
\begin{center}
\includegraphics[width=0.7\linewidth]{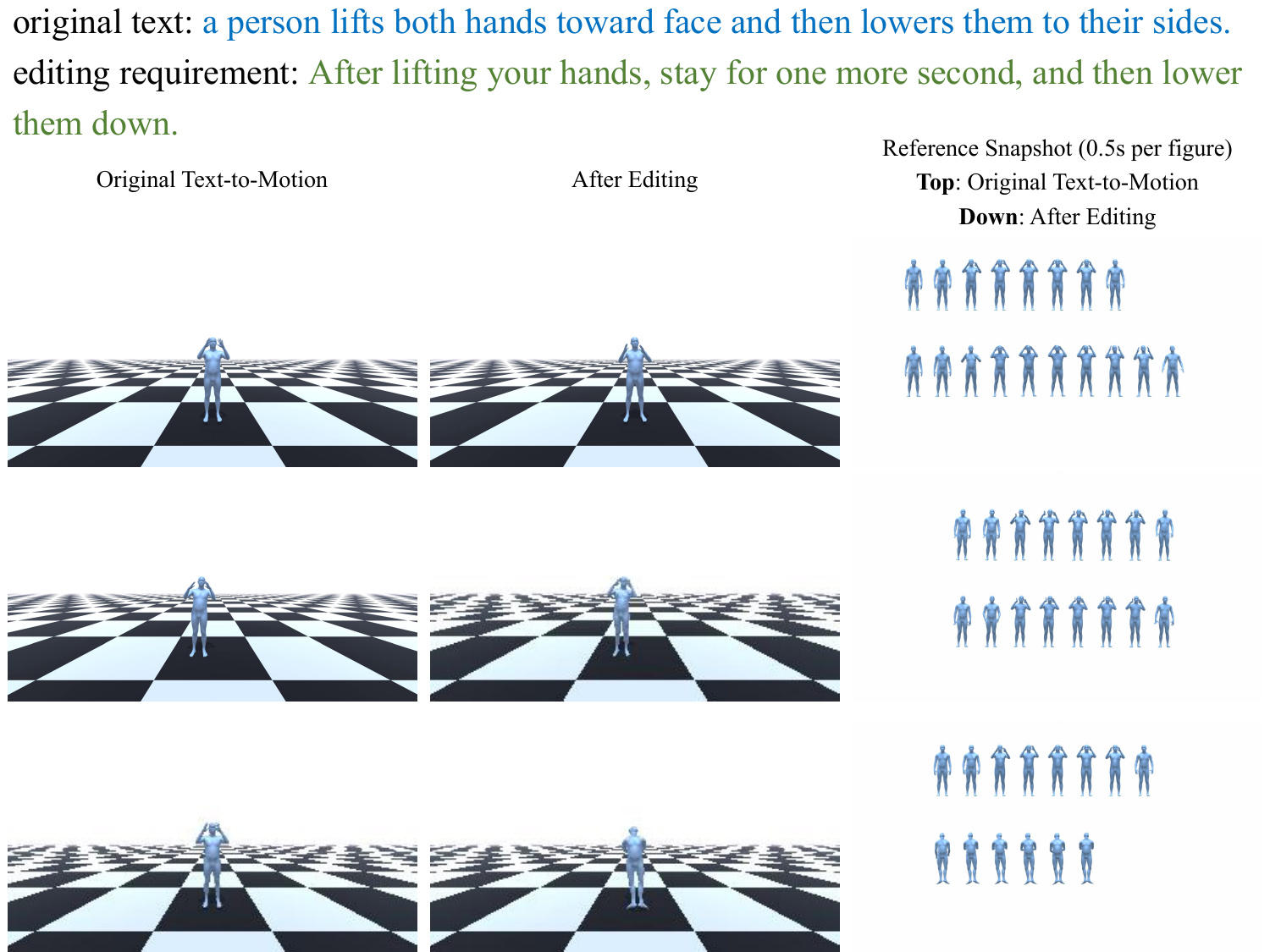}
\end{center}
\end{figure*}

\textcolor{red}{Answer: \quad Row 1: Ours \quad Row 2: FLAME \quad Row 3: T2M-GPT}

\vspace{1em}
\textbf{Case 8}: \textbf{Extend} at the \textbf{end} of the human motion (\textbf{Temporally}).

\begin{figure*}[!h]
\vspace{-1em}
\begin{center}
\includegraphics[width=0.7\linewidth]{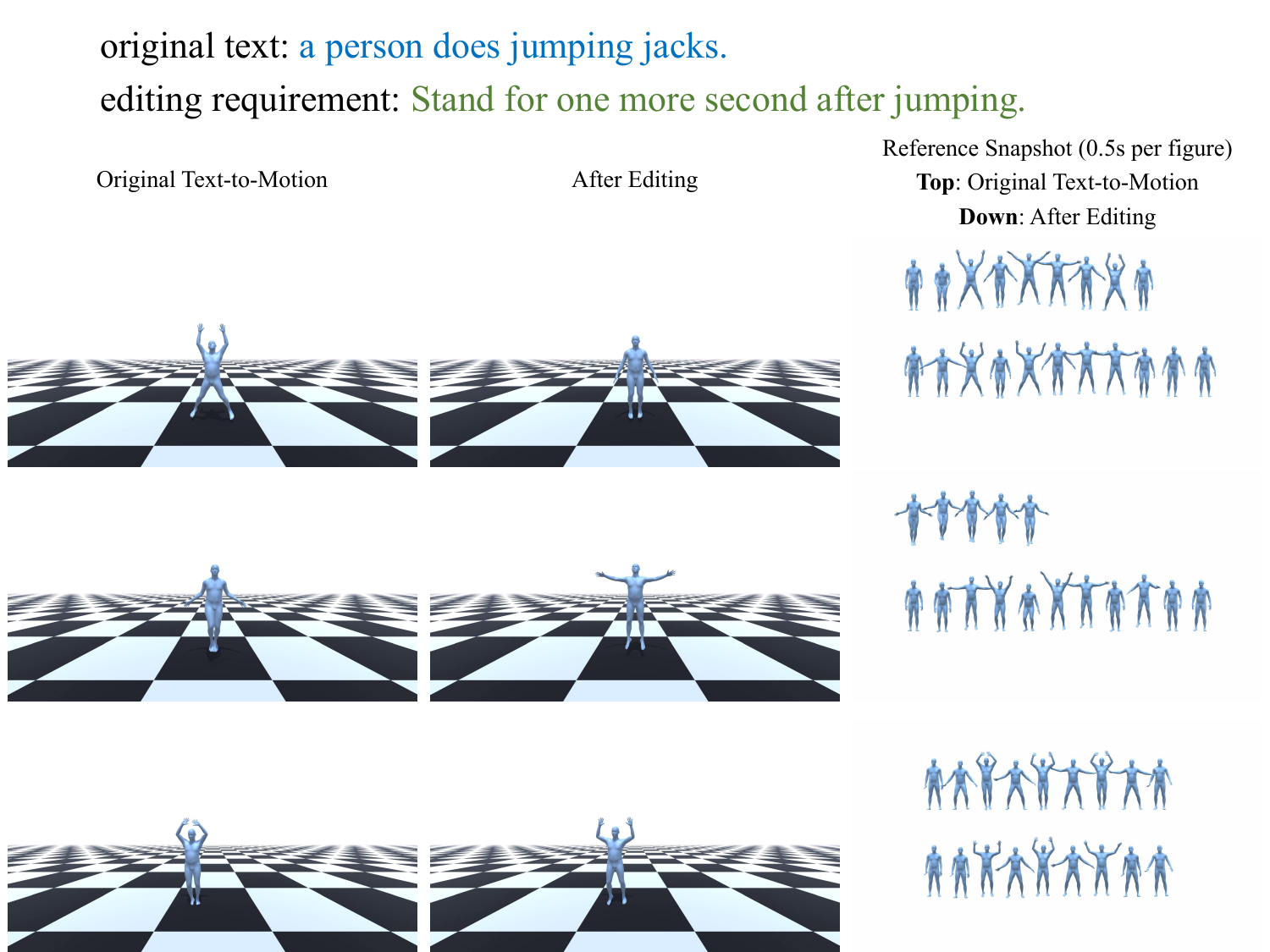}
\end{center}
\end{figure*}

\textcolor{red}{Answer: \quad Row 1: Ours \quad Row 2: T2M-GPT \quad Row 3: FLAME}

\newpage
\textbf{Case 9}: \textbf{Delete} at the \textbf{start} of the human motion (\textbf{Temporally}).

\begin{figure*}[!h]
\vspace{-1em}
\begin{center}
\includegraphics[width=0.7\linewidth]{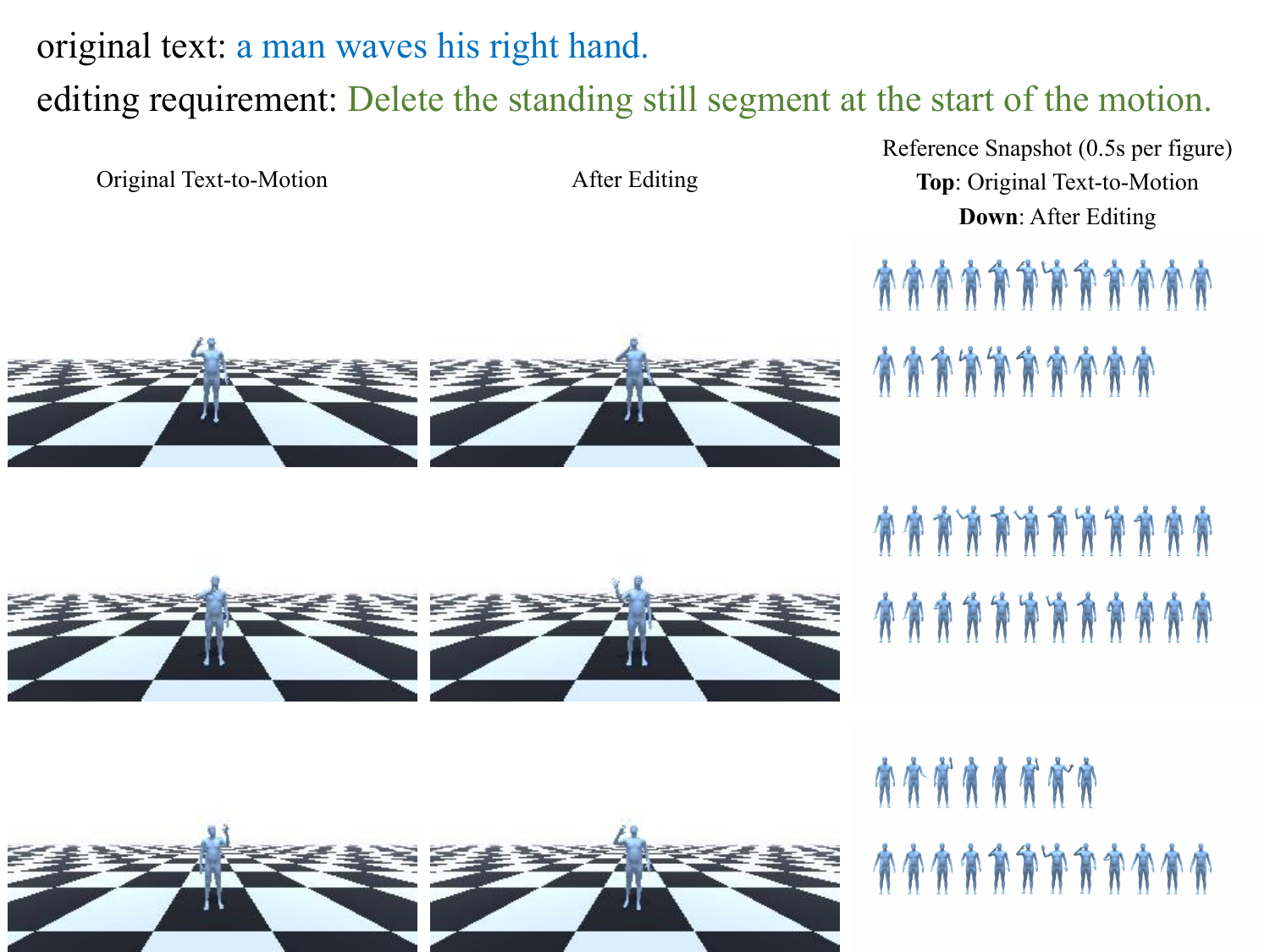}
\end{center}
\end{figure*}

\textcolor{red}{Answer: \quad Row 1: Ours \quad Row 2: FLAME \quad Row 3: T2M-GPT}

\twocolumn
\clearpage
\newpage

\setlength{\floatsep}{5pt} 

\end{document}